\documentclass[10pt,journal,compsoc]{IEEEtran}

\usepackage{amsmath,amsfonts}
\usepackage{savesym}
\savesymbol{Cross}
\usepackage{marvosym}
\restoresymbol{ifsym}{Cross}

\usepackage{savesym}
\savesymbol{Letter}
\usepackage{marvosym}
\usepackage{ifsym}
\restoresymbol{ifsym}{Letter}

\usepackage{algorithm}
\usepackage[noend]{algorithmic}
\usepackage{array}
\usepackage[caption=false,font=normalsize,labelfont=sf,textfont=sf]{subfig}
\usepackage{textcomp}
\usepackage{stfloats}
\usepackage{url}
\usepackage{algorithm}

\usepackage{verbatim}
\usepackage{graphicx}
\usepackage{cite}
\usepackage[utf8]{inputenc} 
\usepackage[T1]{fontenc}    
\usepackage[breaklinks=true,colorlinks,bookmarks=false]{hyperref}
\usepackage{url}            
\usepackage{booktabs}       
\usepackage{amsfonts}       
\usepackage{nicefrac}       
\usepackage{microtype}      
\usepackage{amsthm}
\usepackage{amsmath}
\usepackage{graphicx}
\usepackage{amssymb}
\usepackage{diagbox}
\usepackage{multirow}
\usepackage{pifont} 
\usepackage{bm}
\usepackage{subcaption}
\usepackage{color,xcolor}
\usepackage{caption}
\usepackage{wrapfig}
\usepackage{colortbl}
\usepackage{enumitem}
\usepackage{multirow}
\usepackage{bbding}
\usepackage{pifont}
\usepackage{makecell}
\usepackage{marvosym}

\usepackage[noend]{algorithmic}
\usepackage[utf8]{inputenc} 
\usepackage[T1]{fontenc}    
\usepackage{url}            
\usepackage{booktabs}       
\usepackage{amsfonts}       
\usepackage{nicefrac}       
\usepackage{microtype}      
\usepackage{amsthm}
\usepackage{amsmath}
\usepackage{graphicx} 
\usepackage{amssymb}
\usepackage{diagbox}
\usepackage{multirow}
\usepackage{pifont} 
\usepackage{bm}
\usepackage{subcaption}
\usepackage{color,xcolor}
\usepackage[font=small,labelfont=bf]{caption}  
\usepackage{makecell}
\usepackage{tabulary}
\definecolor{demphcolor}{RGB}{144,144,144}
\newcommand{\demph}[1]{\textcolor{demphcolor}{#1}}
\definecolor{mygray}{gray}{0.4}
\usepackage{colortbl}
\usepackage{enumitem}
\usepackage{multirow}
\usepackage{threeparttable}
\usepackage{multirow}
\usepackage[export]{adjustbox}
\usepackage{threeparttable}
\usepackage{mathrsfs}
\ifCLASSINFOpdf
\else
\fi
\begin{document}

\title{All Nodes are created Not Equal: \\ Node-Specific Layer Aggregation and Filtration for GNN}

       
\author{Shilong Wang, 
        Hao Wu, 
        Yifan Duan, 
        Guibin Zhang, 
        Guohao Li, 
        Yuxuan Liang, 
        Shirui Pan, \\
        Kun Wang\textsuperscript{$*$}, 
        Yang Wang\textsuperscript{$*$} \textit{Senior Member, IEEE}

\IEEEcompsocitemizethanks{\IEEEcompsocthanksitem  Shilong Wang, Hao Wu, Yifan Duan, Kun Wang, Yang Wang are all with University of Science and Technology of China, Hefei, Anhui, P.R.China.

\IEEEcompsocthanksitem Guibin Zhang is with Tongji University and The Hong Kong University of Science and Technology (Guangzhou), P.R.China.

\IEEEcompsocthanksitem Guohao Li is with the University of Oxford.

\IEEEcompsocthanksitem Yuxuan Liang is with INTR Thrust \& DSA Thrust, The Hong Kong University of Science and Technology (Guangzhou). 

\IEEEcompsocthanksitem Shirui Pan is with the Griffith University

\textsuperscript{*} Kun Wang and YangWang are the corresponding authors.

}
}

\markboth{IEEE TRANSACTIONS ON KNOWLEDGE AND DATA ENGINEERING}%
{Wang \MakeLowercase{\textit{et al.}}: Modeling Spatio-temporal Dynamical Systems with Neural Discrete Learning and Levels-of-Experts}

\IEEEtitleabstractindextext{%
\begin{abstract}
The ever-designed Graph Neural Networks (GNNs), though opening a promising path for the modeling of the graph-structure data, unfortunately introduce two daunting obstacles to their deployment on devices. (I) Most of existing GNNs are shallow, due mostly to the \textit{over-smoothing} and \textit{gradient-vanish} problem as they go deeper as convolutional architectures. (II) The vast majority of GNNs adhere to the \textit{homophily assumption}, where the central node and its adjacent nodes share the same label. This assumption often poses challenges for many GNNs working with heterophilic graphs. Addressing the aforementioned issue has become a looming challenge in enhancing the robustness and scalability of GNN applications. In this paper, we take a comprehensive and systematic approach to overcoming the two aforementioned challenges \underline{for the first time}. We propose a \underline{No}de-\underline{S}pecific Layer \underline{A}ggregation and \underline{F}iltration architecture, termed \textbf{NoSAF}, a framework capable of filtering and processing information from each individual nodes. NoSAF introduces the concept of ``All Nodes are Created Not Equal’’ into every layer of deep networks, aiming to provide a reliable information filter for each layer's nodes to sieve out information beneficial for the subsequent layer. By incorporating a dynamically updated codebank, NoSAF dynamically optimizes the optimal information outputted downwards at each layer. This effectively overcomes heterophilic issues and aids in deepening the network. To compensate for the information loss caused by the continuous filtering in NoSAF, we also propose \textbf{NoSAF-D} (Deep), which incorporates a compensation mechanism that replenishes information in every layer of the model, allowing NoSAF to perform meaningful computations even in very deep layers. NoSAF and NoSAF-D have been experimentally verified across various graph benchmarks (11 homophilic and 6 heterophilic datasets). The results demonstrate that NoSAF and NoSAF-D surpass SOTAs ($9$ methods) across all shallow and deep architectures, as well as in both homophilic and heterophilic scenarios. \textbf{\textit{Code is available in \url{https://github.com/wslong20/NoSAF-NoSAF-D}.}}

\end{abstract}

}

\maketitle

\IEEEdisplaynontitleabstractindextext

%
\IEEEpeerreviewmaketitle

\section{Introduction}
\label{sec:introduction}

Graph Neural Networks (GNNs) have achieved significant success in various graph representation learning tasks, including node classification \cite{wang2019dynamic, hamilton2017representation, chiang2019cluster}, link prediction \cite{kipf2016variational, zhang2018link}, and graph classification \cite{zhang2018end, gilmer2017neural}. The plausible reason is that, GNNs typically employ a message passing or neighborhood aggregations mechanism \cite{wu2020comprehensive, zhou2020graph}, which iteratively \textit{aggregates} information from neighboring nodes and \textit{updates} the central node to form more \textit{\underline{discriminative}} node representations. Consequently, the final node representations blend in the structural information, significantly improving the model's predictive performance \cite{li2020deepergcn, li2019deepgcns, zhang2024graph, wang2023brave, zhang2024heads, fang2024exgc}. 

\begin{figure}[t]
\centering
\includegraphics[width=1.0\columnwidth]{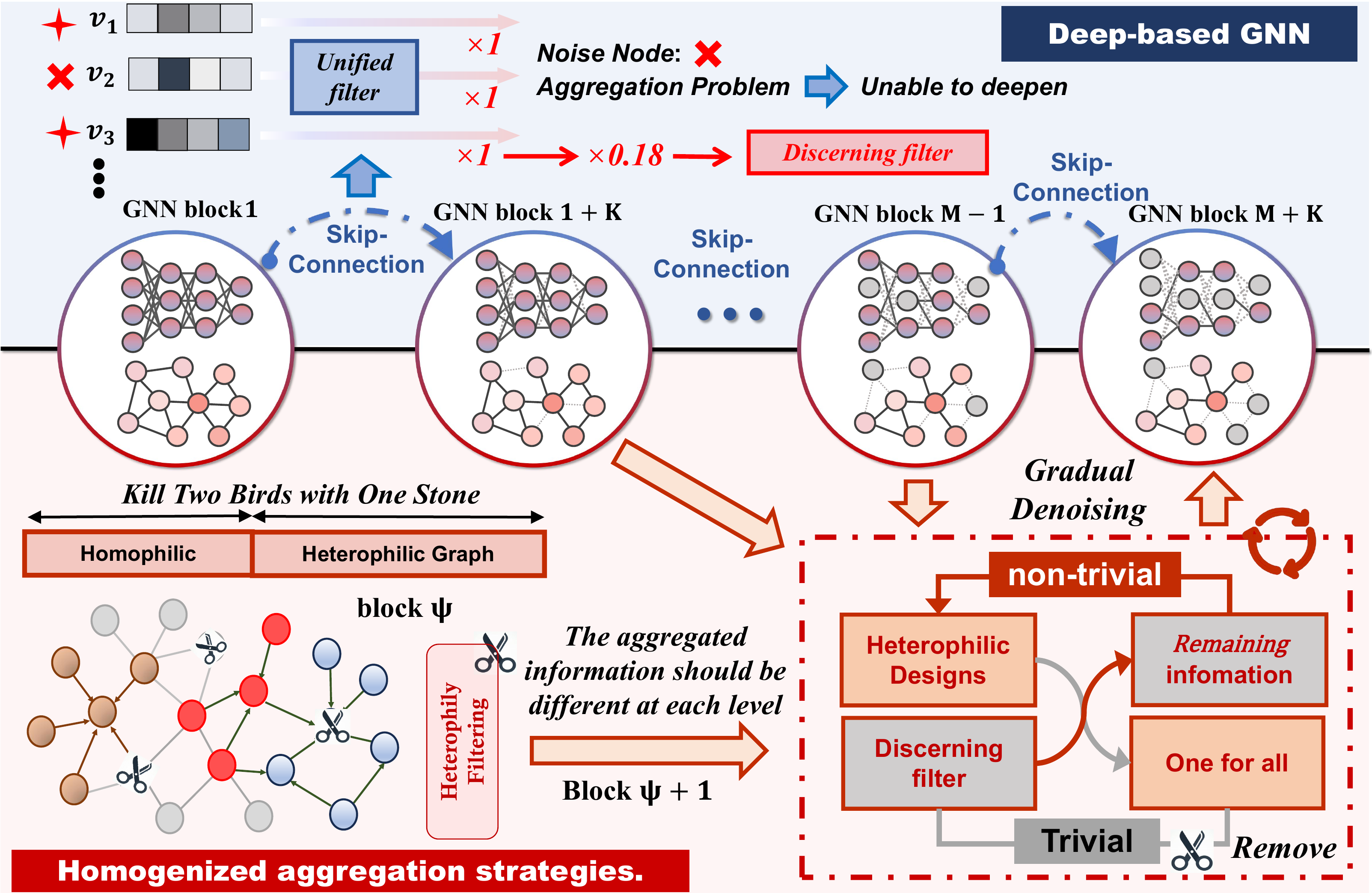}
\vspace{-0.4cm}
\caption{Challenges in deep GNN architectures (${\mathcal D}1\sim{\mathcal D}1$): limited node discrimination with skip-connections (the upper pipeline) and inadequate handling of heterophilic graphs (lower section).}
 \vspace{-0.3cm}
\label{fig:intro}
\end{figure}

Despite their remarkable achievements, the majority of existing GNNs are characterized by shallow architecture \cite{li2018deeper, feng2021should}, primarily because deepening these networks often leads to over-smoothing \cite{keriven2022not, chen2020measuring} and gradient vanishing \cite{li2020deepergcn} issues. With SOTAs like GCN and GAT reaching their peak performance with only $2\sim4$ layers. As one of the most well-established approaches in deepening the structure of GNNs, deep GNN appears to lag behind in the era of CV, due mostly to its cumbersome message-passing designs \cite{abu2019mixhop, rong2019dropedge, cui2023mgnn} and the reckless adoption of CV designs like skip connections \cite{li2019deepgcns, xu2018representation, chen2020simple, li2020deepergcn}, these approaches sadly place a daunting obstacle in the path of adapting these models effectively for on-graph applications. 


Upon close examination of the current mainstream deep GNN models, they can be primarily categorized into three distinct types: \textcolor{black}{\ding{182}} the first class \cite{li2019deepgcns, li2020deepergcn, chen2020simple, xu2018representation, huang2021towards} draw inspiration from the field of computer vision, incorporating techniques such as skip connections or cross-layer information fusion \cite{he2016deep, huang2017densely, ronneberger2015u, long2015fully}. In line with the operational mechanism of GNNs, they enhance the final node representations by integrating information from the shallower layers of the network, effectively mitigating the over-smoothing problem. These deep GNN designs have achieved optimal performance in a variety of graph-related tasks \cite{li2019deepgcns, li2021training, chen2020simple}. \textcolor{black}{\ding{183}} The second class \cite{gasteiger2018predict, liu2020towards, frasca2020sign, wu2019simplifying} separate neighborhood aggregation process from the update process (\textit{typically categorized as feed-forward layer}), performing it before or after update layer, \textit{e.g.,} APPNP \cite{gasteiger2018predict}. Based on Personalized PageRank \cite{lofgren2016personalized}, APPNP limits the size of the local receptive field, maintaining local information even with continuous neighborhood aggregation, countering over-smoothing. \textcolor{black}{\ding{184}} Another research line \cite{rong2019dropedge, feng2020graph} don't change the network architecture but implicitly introduce data augmentation techniques into the training process, like DropEdge \cite{rong2019dropedge}, which randomly samples the adjacency matrix during training, reducing node connections and increasing data diversity to alleviate over-smoothing and overfitting.

Notably, although these three categories of deep GNN frameworks achieve higher accuracy than simpler ones, they inevitably encounter certain drawbacks ($\mathcal{D}$) as detailed below:

\begin{itemize}[leftmargin=*]
    \item[{${\mathcal D}1$}:] Recklessly consolidating shallow-layer information into a ``one-size-fits-all" package for deeper layers may lead to a loss of nuanced insights into the individual nodes. In reality, each node in the shallow layers contributes differently to the information processed by the deeper layers \cite{anonymous2023graph, wang2023snowflake}. From a vertical perspective (considering the network's depth \textcolor{black}{\ding{182}}), it's necessary to make discerning evaluations of the outputs from each node.

    \item[{${\mathcal D}2$}:] \textbf{Homogenized aggregation strategies.} As depicted above, customized aggregation strategies often introduce additional auxiliary modules and intricate designs to enhance their performance. However, due to their complex design, these strategies are neither simple nor universal, lacking insights for various tasks or datasets. More critically, as different nodes share a uniform aggregation strategy (the \textcolor{black}{\ding{183}} and \textcolor{black}{\ding{184}} research line) \cite{gasteiger2018predict}, the performance of current GNNs has sharply declined with the decrease in graph homophily ratio. Specifically, within heterophilic graphs, there's often a discrepancy observed between the labels of neighboring nodes and the central node \cite{yang2016revisiting, bojchevski2017deep, shchur2018pitfalls, mernyei2007wikipedia}. This scenario may necessitate the implementation of distinct processing approaches for each neighboring node \cite{chen2020measuring, zheng2022graph, luan2022revisiting, wang2023snowflake}.
\end{itemize}

\vspace{-0.5em}
\noindent Learning from the lessons outlined above, an effective strategy for \textbf{\textit{cross-layer information fusion}}, along with an approach which capable of \textbf{\textit{discerning the effectiveness of information aggregated from local neighborhoods,}} are both crucial for the training of graph models. Toward this end, we propose a \underline{No}de-\underline{S}pecific Layer \underline{A}ggregation and \underline{F}iltration archietecture, termed as \textbf{NoSAF}, a framework that capable of filtering and processing information from different nodes. NoSAF dynamically maintains a codebank \cite{van2017neural,mohamed2022self}, which is adept at handling each node distinctly, determining the proportion of information to be retained for every individual node. In the process of filtering from shallow to deeper layers, \textit{NoSAF governs the codebank to extract information at each level and subsequently updates the codebank}. Ultimately, the output from the codebank is utilized as the model's output.

NoSAF eschews the introduction of complicated multi-branch designs (\textit{e.g. branch concatenation in Inception}), which often complicate implementation and customization. The codebank is both simple and efficient, adding a small storage overhead—yet it significantly enhances the model's performance. Moreover, NoSAF's customized evaluation for each node makes it \textbf{\textit{inherently suited for addressing heterophilic graph problems}}. 


\textbf{\ding{224} Improvement.} Due to the instability of information filtering through layers, we also present a version of NoSAF adapted for deep GNNs, termed NoSAF-D. This version integrates the codebank into the main network to prevent the collapse of information in intermediate layers due to continuous information filtering, ensuring that the network remains stable and robust even when training ultra-deep models.\textbf{Contributions} can be summarized as follows:

\begin{itemize}[leftmargin=*]
    \item We propose NoSAF, a straightforward and resilient codebank architecture designed to deepen GNN networks and adeptly handle both heterophilic and homophilic network designs.
    
    \item Codebank introduce an universal way for learning varying importance levels in node representations, significantly reducing the noise in post-fusion information. Further, the concept of codebank is model-agnostic, which can be seamlessly transfer to various GNNs like GCN \cite{kipf2016semi}, GAT \cite{velivckovic2017graph}, and others \cite{hamilton2017inductive}.
    
    \item We conducted extensive experiments to validate the capability of NoSAF, including its application on 17 benchmark datasets, which encompass both homophilic and heterophilic graphs. Our experiments consistently demonstrate that the proposed approach outperforms competitors, particularly in scenarios involving deep networks and varying degrees of heterophily.
\end{itemize}

\section{Related Work}
Our work builds upon the following research foundations:

\textbf{Neighborhood Aggregation \& message passing Schemes} are the  key factors in improving the performance of GNNs. The distinction between various GNNs often stems from their unique neighborhood aggregation scheme. GCN and GCN-like \cite{kipf2016semi, velivckovic2017graph, gilmer2017neural} models typically update node features by weighted summation of features from one-hop neighbors and the node itself. Graph Diffusion Networks (GDNs) \cite{gasteiger2019diffusion, sun2020adaptive, gasteiger2018predict} extend aggregation to multi-hop neighbors, assigning lower weights to distant neighbors. This approach retains more local information while also enabling nodes to perceive distant neighbors, thus enhancing performance. However, extending the neighborhood to multi-hops often involves power operations on adjacency matrix, which are computationally complex. Some works \cite{gasteiger2018predict, zhu2020simple, wu2019simplifying} that separate neighborhood aggregation from networks can be considered as specific case of GDNs \cite{gasteiger2018predict}.

\textbf{Deep GNNs.} Drawing from the realm of CV, Deep GNNs incorporate skip or residual connections \cite{he2016deep,huang2017densely} to allow cross-layer information fusion. This enables GNNs to integrate node representations learned at shallow layers (local information) into deeper layers, effectively addressing the issue of over-smoothing. The JKNet \cite{xu2018representation} introduces Jumping knowledge connections, enabling the model to adaptively sum the intermediate representations. DeepGCN \cite{li2019deepgcns} and its variant \cite{li2019deepgcns,li2020deepergcn} introduce residual and dense connections that add shallow-layer representations directly to deep-layer representations. GCNII \cite{chen2020simple} combines two types of residuals, incorporating a portion of the model's input into all subsequent layers, thus preserving some initial information in each layer's input.

\textbf{Heterophilic GNNs.} In the real world, many heterophilic graphs \cite{pei2020geom,rozemberczki2021multi} exist, which the majority of neighbors connected to each node often have labels that differ from the node itself. Consequently, GNNs predicated on the assumption of homophily do not perform well on these datasets. Consequently, much effort has been dedicated to enhancing the performance of GNNs on heterophilic graphs \cite{wu2020comprehensive, zhu2021graph}. Geom-GCN \cite{pei2020geom} aggregates neighbor information by mapping nodes into a latent geometric space and utilizing geometric relations in this space.  MixHop \cite{abu2019mixhop} operates by simultaneously aggregating information from multi-hop neighbors at each layer. This allows the model to capture not only local neighborhood features but also understand broader graph structures.

\section{Preliminaries}

\subsection{Graph Definition}
A graph $\mathcal{G}$ can be defined as a tuple $\mathcal{G} = \left(\mathcal{V}, \mathcal{E} \right)$, where $\mathcal{V}$ denotes the set of nodes, and $\mathcal{E}$ is the set of edges representing the connectivity between nodes $v \in V$. An edge $e_{i,j} \in \mathcal{E}$ implies that there is a connection between the nodes $v_i$ and $v_j$. In a graph, each node $v_i$ is associated with a feature vector $x \in \mathbb{R}^{D}$, where $D$ is the feature dimension. The feature vectors of all vertices in the graph collectively form the graph's feature matrix $\mathbf{X} \in \mathbb{R}^{N \times D}$, where $N$ represents the number of nodes in the graph.

\noindent \textbf{Neighborhood Aggregation.} GNNs typically update node features at each layer by aggregating the features of neighbor nodes. Let's define the feature of node $v_i$ at the $l$-th layer as $h_{i}^{l} \in \mathbb{R}^{d}$, and the feature matrix for the $l$-th layer as $H^l \in \mathbb{R}^{N \times d}$, where $d$ represents the dimension of the node features at layer $l$. For simplicity, we assume that $d$ is the same across all layers. The neighborhood of node $v_i$ is denoted as $N\left( v_i \right)$, and specifically for a first-order neighborhood, it can be defined as $N(v_i) = \{ v_j \in \mathcal{V}|\left( v_i, v_j \right) \in \mathcal{E} \}$.

 The neighborhood aggregation scheme at the $l$-th layer of GNNs can be written as follows:
 \begin{equation}
 h_{i}^{(l)} = \sigma \left( W^l \cdot \text{Aggregate} \left( \{ h_j^{(l-1)} | v_j \in \mathcal{N}(v_i) \} \right) \right)
 \end{equation}
where \textbf{Aggregate} represents an aggregation function defined by the specific model. $W^l$ is the trainable weight parameter of the $l$-th layer. $\sigma$ denotes a nonlinear activation function, typically ReLU.

\subsection{Homophily and Heterophily Ratio}

Following previous work \cite{pei2020geom}, we define the node- and graph-level homophily ratio as follows:
\begin{equation}
     \quad \mathcal{H}_{i} = \frac{|\{v_j| v_j \in \mathcal{N}(v_i), \;y_i = y_j\}|}{|\mathcal{N}(v_i)|}  ,
\end{equation}
\begin{equation}
    \quad \mathcal{H} = \frac{1}{N} \sum_{i=1}^N \mathcal{H}_{i},
\end{equation}
$\mathcal{H}_{i}$ represents the proportion of neighbors that share the same class with node $v_i$, and $\mathcal{H}$ represents the global homophily by computing the average of node homophily. Conversely, the heterophily ratio at the node and graph level can be expressed as $\widetilde{\mathcal{H}}_{i} = 1 - \mathcal{H}_{i}$ and $\widetilde{\mathcal{H}} = 1 - \mathcal{H}$. In general, graphs that exhibit strong homophily tend to have $\mathcal{H}$ values approaching 1, whereas those characterized by pronounced heterophily often have values near 0.



\section{Motivation}\label{sec:motivation}
\begin{figure}[t]
\centering
\includegraphics[width=1.0\columnwidth]{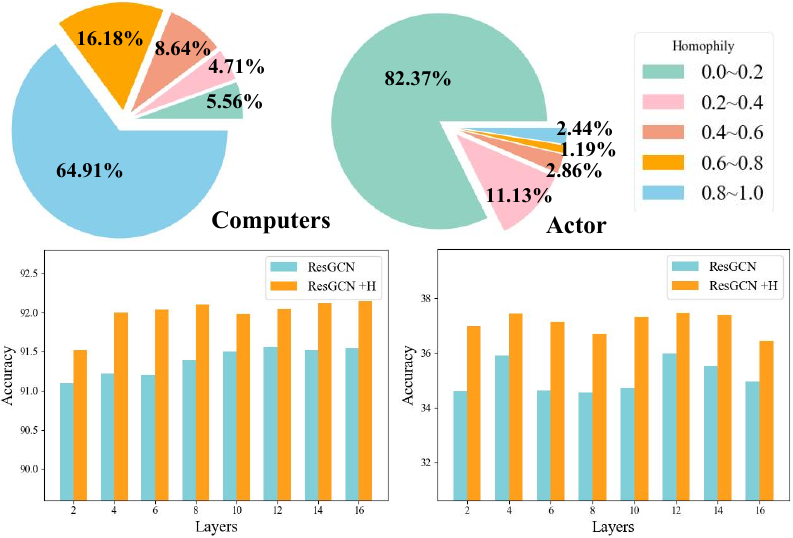}
\vspace{-0.6cm}
\caption{The motivation of our NoSAF framework.The pie charts depict the proportion of nodes with varying homophily in the Computers dataset ($\mathcal{H}=0.79$) and the Actor dataset ($\mathcal{H}=0.22$). In the bar charts, "+H" indicates that the model assigns $\mathcal{H}_i$ as the weight for aggregating information for each node $i$.}
 \vspace{-0.3cm}
\label{fig:motivation}
\end{figure}

In this section, we present a straightforward motivation for our proposed NoSAF. As shown in Fig \ref{fig:motivation}, from the two upper charts in Figure 2, it is apparent that for the homophilic graph Computers ($\mathcal{H}=0.79$), 64.91\% of nodes have a homophily value between $0.8\sim1.0$. Conversely, in the heterophilic graph Actor ($\mathcal{H}=0.22)$, 82.37\% of the nodes have over 80\% of their adjacent nodes with a different label. Therefore, when a neighborhood aggregation strategy is applied to a homophilic graph, most nodes update their features through the characteristics of adjacent nodes with the same label. This leads to increasingly similar representations among nodes with the same label, which is highly beneficial for node classification tasks. On the other hand, when applied to a heterophilic graph, the aggregated information from neighbors may actually disrupt the node's inherent features, leading to a decline in model performance.

We conjecture that \textit{assigning a certain weight to the features of each node after aggregation to gauge their importance might enhance the model's capability.} To test this conjecture, we design a simple experiment: we use ResGCN \cite{li2018deeper} as backbone and apply the homophily value of each node as the weight for its aggregated features at every layer. This approach allows the features aggregated from nodes with high homophily to receive higher weights, while those from nodes with low homophily rates receive lower weights. The experimental results demonstrate that differentiating the aggregated information of nodes can significantly improve the model's performance, both on homophilic and heterophilic graph datasets. However, the calculation of homophily value relies on the labels of nodes. Since the labels of nodes in the validation and test sets cannot be utilized during the inference stage, the homophily value cannot be employed as a basis for assessing the importance of nodes in practical training. Furthermore, using homophily value to evaluate the quality of each node's aggregated information does not allow the model to discern finer details within the graph. For instance, even if two nodes share the same homophily value, their local structures can still differ. Consequently, we propose NoSAF, a method where each layer is designed to automatically learn these weights. This capability allows NoSAF to effectively address the complexities inherent in more intricate graph structures.

\section{Methodology}
\begin{figure*}[t]
    \centering
    \includegraphics[width=1.00\linewidth]{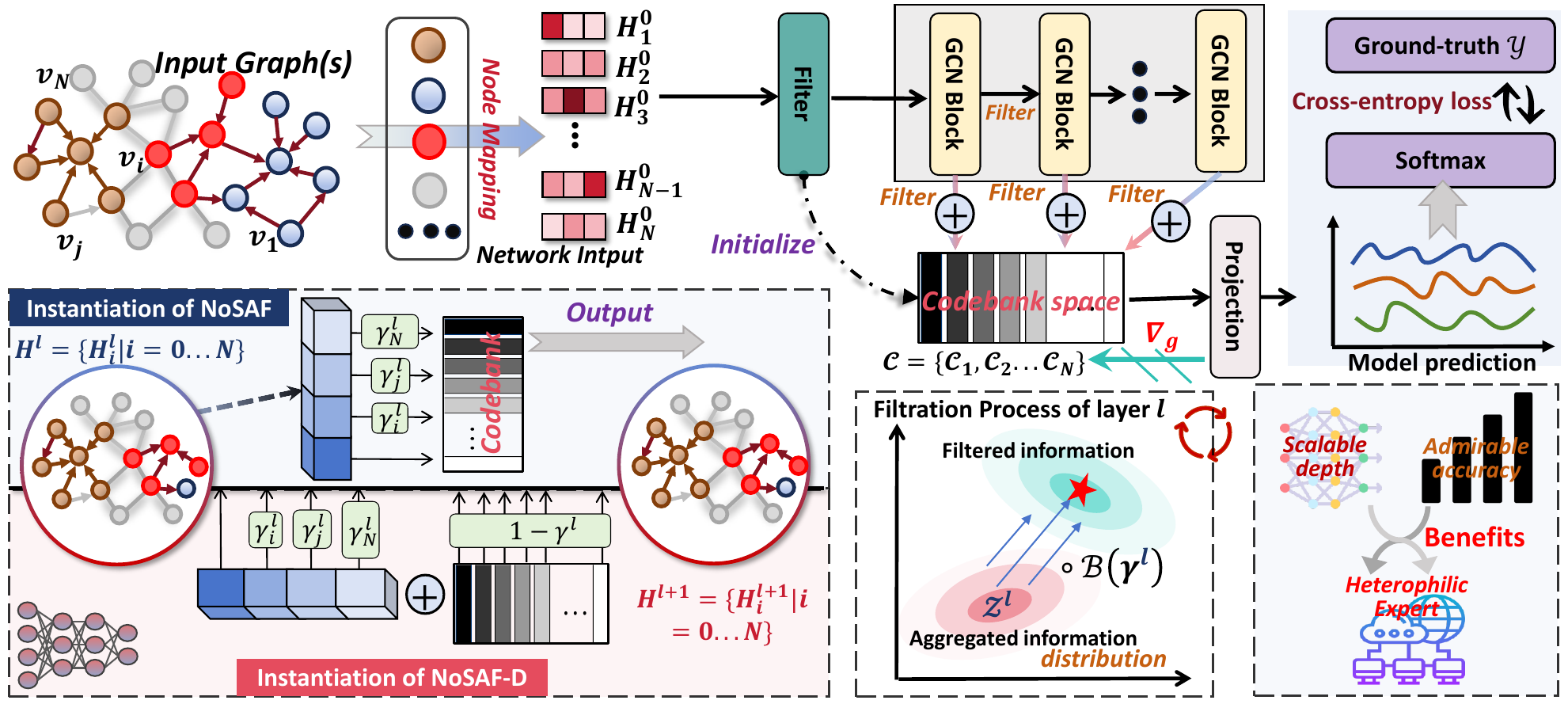}
    \caption{The model overview of NoSAF and NoSAF-D version. NoSAF assigns weight ${\gamma}^l_{i}$ to each node $i$ at layer $l$, implementing hierarchical filtering. It also dynamically updates the codebank to ensure that the most optimal information is retained for output. "Filter" refers to the filtration operation as demonstrated by Equation (9).}
    \label{fig:model}
\end{figure*}

\noindent \textbf{\textit{Overview.}} We scale up GCN model with our NoSAF concept in this section. The NoSAF framework overview is illustrated in Figure \ref{fig:model}. Our design encompasses two critical directions. The NoSAF version adeptly addresses the heterophilic graph issue, while NoSAF-D is further tailor-made to enhance the depth of GNN layers.

\subsection{The Basic Pipeline of NoSAF}

As discussed in \cite{chen2020simple}, the GCN model employs a message-passing mechanism to assist the central node in aggregating connected neighboring nodes. This process updates the node to form a more discriminative node representation. Initially, the model's input $X\in \mathbb{R}^{N \times D}$ is fed into an mapping layer $\zeta_{in}$ (typically MLP), which then mapping into hidden representation $H^{0} \in \mathbb{R}^{N \times d}$:
\vspace{-0.5em}
\begin{equation}
    H^{0} = \zeta_{in} \left( X \right) \in \mathbb{R}^{N \times d}
\end{equation}

\noindent The encoded feature $H^{0}$ serves as the input for the GNN backbone, and the input dimension for each subsequent layer of the GNN backbone is uniformly $d$. It is worth mentioning that although we employ GCN as the backbone network, our concept is model-agnostic. We utilize GCN as a general framework to better illustrate the effectiveness of our algorithm.

Given our work's focus on deepening GNN networks, we combine GCN convolution with Batch Normalization (BN) to form a GCN block. BN aids in faster network convergence as it mitigates the effect of internal covariate Shift.
\begin{equation}
    \widetilde{D}_{ii} = \sum_{j=1}^{N} \widetilde{A}_{ij},  \quad \xi_{i}^{l} = \sum_{\substack{j \in \{N(i) \cup i\}}} \frac{\widetilde{A}_{ij}}{\sqrt{\widetilde{D}_{ii}\widetilde{D}_{jj}}} H_{j}^{l}W^{l}
\end{equation}
\noindent The above equation describes the aggregation process of a GCN block. where $A$ denotes the adjacency matrix, $\widetilde{A}=A+I$ represents the self-looped adjacency matrix, $\widetilde{D}$ represents the diagonal degree matrix. GCN updates itself through the expressions of its connected neighbors $\{N(i) \cup i\}$. After that, GCN block employs BN)and the nonlinear activation function ReLU, denoted as $\sigma$, to shape the output expression of each layer:

\begin{equation}
    \quad \mathcal{Z}^{l} = \sigma\left({\rm{BN}}\left(\left[\xi_{0}^{l}, \xi_{1}^{l}, \ldots, \xi_{N}^{l}  \right]\right)\right)
\end{equation}

\noindent $\mathcal{Z}^{l}$ represents the complete expression of all nodes at the $l$-th layer. Going beyond this process, we implement two \underline{modifications} to ensure that the NoSAF can adeptly address both the challenges of heterophilic graphs and the need for deeper network layers, effectively killing two birds with one stone.

\subsection{One Node One Filter Weight}

As depicted in Sec. \ref{fig:motivation}, each node contributes differently to deeper layers of the network. This is especially true in heterophilic graphs, where employing a uniform method to feed node expressions into deeper layers may deteriorate the final output expression. To this end, we introduce a dynamic filter for each node, designed to learn the specific information that should be retained for each node, termed as \textbf{node weights} (\textbf{Modification} \textcolor{black}{\ding{182}}). To better preserve the updated state of each node at every layer, we introduce a dynamically updated codebank, which provides a powerful prior expression for each layer as it progresses to deeper levels (\textbf{Modification} \textcolor{black}{\ding{183}}).

Specifically, given the Codebank $\mathcal{C}^{l}$ and the node features $\mathcal{Z}^{l}$ aggregated at the $l^{th}$ layer, we utilize the values in the codebank to gauge the importance of aggregated information.

\vspace{-0.3em}
\begin{equation}
\begin{aligned}
     \quad \beta_{i}^{l} = \mathcal{Z}_{i}^{l}W_{\mathcal{Z}}^{l} \| \mathcal{C}_{i}^{l}W_{\mathcal{C}}^{l} \;\;\; \in \;\; {\beta ^l} = {\rm{set}}\left\{ {\beta _1^l,\beta _2^l, \ldots ,\beta _N^l} \right\}
\end{aligned}    
\end{equation}
\vspace{-0.3em}
where $W_{\mathcal{Z}}^{l}, W_{\mathcal{C}}^{l} \in \mathbb{R}^{d \times d'}$, $\|$ represent the concatenation operation. At the $0$-th layer, $\mathcal{Z}^{0}=H^{0}$, $\mathcal{C}^{0}={\textbf{0}^{N \times d}}$. $\beta_{i}^{l} \in \mathbb{R}^{2d'}$, which is a hybrid variable that includes information from the codebank as well as the aggregated information of the node. Unlike previous works that focus on the importance of connected nodes \cite{wang2021towards, luo2020parameterized}, our approach \textbf{\underline{first}} concentrates on the significance of nodes across different layers. With the hybrid variable $\beta_{i}^{l}$ at hand, we generate weight for node $i$ as following:
\vspace{-0.3em}
\begin{equation}
    \quad \gamma_{i}^{l} = {\rm{Sigmoid}}\left({\rm{LeakyReLU}}_{\lambda}\left( \beta_{i}^{l}W_{1}^{l} + b_{1, i}^{l} \right)W_{2}^{l} + b_{2, i}^{l}\right)
\end{equation}
\vspace{-0.3em}
where $W_{1}^{l} \in \mathbb{R}^{2d' \times d''}$ and $W_{2}^{l} \in \mathbb{R}^{d'' \times 1}$ are, respectively, the weight matrices for the first and second linear transformations. We employ LeakyReLU with a negative slope of $\lambda$ as the nonlinear activation function. Finally, we generate the importance score $\gamma_{i}^{l}\in \left(0, 1\right)$ for node $i$. By assigning the corresponding importance score as filter weight to each node, we can dynamically update each node's contribution to the deeper layers of the network. We denote this filtering operation as $\mathcal{F}_f$:

\begin{equation}
    {{\mathcal F}_{f}}\left( {{{\mathcal Z}^l},\gamma^l } \right) = {{\mathcal Z}^l} \circ {\mathcal B}\left( {{{\rm{\gamma }}^l}} \right) \;\; {\rm{for}} \;\;\;\; l=1,2...L
\end{equation}

\noindent We iteratively update the filter for each layer to ensure that each one effectively filters out information that is not useful for deeper layers. ${\mathcal B}( {{{\rm{\gamma }}^l}} )$ operator broadcast $\gamma^{l} \in  \mathbb{R}^{N}$ into $\mathbb{R}^{N \times d}$ dimensions for Hadamard product ($\circ$) with $\mathcal{Z}^{l}$. Consequently, each row of ${\mathcal B}( {{{\rm{\gamma }}^l}})$ contains identical values. Based on this, we define the input information at $l+1$-th GCN block $H^{l+1}$ as $H^{l+1} = {{\mathcal F}_f}( {{{\mathcal Z}^l},\gamma^l } )$.

\textbf{\textit{Codebank Updating.}} The filtering mechanism applies a weight between 0 and 1 to the aggregated information of the nodes, reducing the amount of noise accumulated in the codebank and diminishing many meaningless aggregations in subsequent layers caused by the disruption of node features. Subsequently, we introduce a dynamically updated Codebank to iteratively refine the information, ensuring the final update yields the optimal output. Assume that the update state of the codebank at the $l^{th}$ layer is denoted as $\mathcal{C}^{l}$. We summarize the layer-wise update formula as follows:

\begin{equation}
    \quad \mathcal{C}^{l+1} = \mathcal{C}^{l} +{{\mathcal Z}^l} \circ {\mathcal B}\left( {{{\rm{\gamma }}^l}} \right) \;\; {\rm{for}} \;\;\;\; l=0,1,2...L
\end{equation}

\noindent We use the information in codebank from the last layer as the representation learned by the backbone network. We summarize the update process in following equation:

\begin{equation}
\begin{split}
    \quad \mathcal{C}^{L} &= \mathcal{Z}^{L} \circ {\mathcal B}( {{{\rm{\gamma }}^L}}) + \mathcal{C}^{L-1} \\
                    &= \mathcal{Z}^{L} \circ {\mathcal B}( {{{\rm{\gamma }}^L}}) + \mathcal{Z}^{L-1} \circ {\mathcal B}( {{{\rm{\gamma }}^{(L-1)}}}) + \mathcal{C}^{L-2} \\
                    &= \ldots \\
                    &= \sum_{l=0}^{L} \mathcal{Z}^{i} \circ {\mathcal B}( {{{\rm{\gamma }}^i}})  \\ 
                    &= \sum_{l=0}^{L} {{\mathcal F}_f}( {{{\mathcal Z}^l},\gamma^l } )
\end{split}
\end{equation}

\noindent Therefore, the final updated code repository can be represented as the sum of filtered information across all layers. Finally, we use an mapping layer $\zeta_{out}$ to extract features from $\mathcal{C}^{L}$ and map them to the dimension $K$, where $K$ represents the number of classes for classification. The model is then trained using a cross-entropy loss ${\mathcal{L}}_{\eta,y}$ function:

\vspace{10pt}
\begin{minipage}{.461\textwidth}
\begin{equation}
    \quad \eta = \zeta_{out}\left(\mathcal{C}^{L}\right),
\end{equation}
\end{minipage}%

\begin{minipage}{.461\textwidth}
\begin{equation}
    {\mathcal{L}}_{\eta,y} = -\frac{1}{N} \sum_{i=1}^{N} \log \frac{\exp(\eta_{i,y_i})}{\sum_{c=1}^{K} \exp(\eta_{i,c})}
\end{equation}
\end{minipage}
\vspace{5pt}


\noindent \textbf{Comparison with Skip-connection.} \textit{Existing DeepGNN architectures \cite{li2019deepgcns, li2020deepergcn, chen2020simple, xu2018representation} often aim to deepen their structure by incorporating shallow, over-smoothed node expressions to mitigate the over-smoothing issue, achieving significant results. However, it is crucial to recognize that recklessly transplanting techniques from computer vision to GNNs inevitably has drawbacks. The issue of heterophily in graphs requires each node to consider the consistency between its own expression and that of its neighbors during aggregation. This leads to concerning performances in heterophilic graphs for identity mapping products like GCNII and residual works like JKNet and ResGCN. Deepening the network and addressing heterophily are unavoidable challenges in GNN design. Our work overcomes these issues elegantly with the codebank method, \textbf{simultaneously filtering importance for each node and addressing both deepening and heterophily}}.

\subsection{The Deepening Objective (NoSAF-D)}

\begin{algorithm} 
\caption{The framework of NoSAF and NoSAF-D}
\label{alg:NoSAF}
\begin{algorithmic}[1]
\STATE \textbf{Input:} $\rm{\mathcal{D}} = \left(\rm{X}, \rm{Y}, \rm{A}\right)$. Network hidden dimension $d$, filter hidden dimension $d'$, network layers $L$. and the total number of training epochs $T$.

\STATE \textbf{Initialize} $\mathcal{C}^{0}$ and network weights.

\FOR{$t=0, 2,...,T-1$}
    \FOR{$l=0,1,...,L$}
        \IF{$l = 0$}
            \STATE Compute $\mathcal{Z}^{0} = H^{0} = \zeta_{in} ( X )$
        \ELSE
            \STATE Compute $\mathcal{Z}^{l}=\rm{BN}(\rm{Aggregate}(H^{l}))$
        \ENDIF 
        \STATE Concatenate $\mathcal{Z}^{l}, \mathcal{C}^{l}$ get $\beta^{l}$
        \STATE Compute $\gamma^{l} = {\rm{Sigmoid}}({\rm{LeakyReLU}}_{\lambda}( \beta^{l}W_{1}^{l} + b^{l} )W_{2}^{l} + b^{l})$
        \STATE Let $\mathcal{F}_f(\mathcal{Z^l, \gamma^l}) = \mathcal{Z}^{l} \circ \mathcal{B}(\gamma^{l})$
        \STATE Let $\mathcal{F}_{cpm}(\mathcal{C}^l, \gamma^l)=\mathcal{C}^{l} \circ \mathcal{B}(1 - \gamma^{l})$
        \IF{NoSAF}
            \STATE Compute $H^{l+1} = \mathcal{F}^{l}(\mathcal{Z, \gamma})$
        \ELSIF{NoSAF-D}
            \STATE Compute $H^{l+1} = \mathcal{F}^{l}(\mathcal{Z, \gamma}) + \mathcal{L}_{cpm}^{l}(\mathcal{C}, \gamma)$
        \ENDIF
        \STATE Compute $\mathcal{C}^{l+1}=\mathcal{C}^{l} + \mathcal{F}^{l}(\mathcal{Z, \gamma})$
    \ENDFOR
    \STATE Compute $\eta = \zeta_{out}( C^{L} )$
    \STATE Compute loss ${\mathcal{L}}_{\eta,y} = -\frac{1}{N} \sum_{i=1}^{N} \log \frac{\exp(\eta_{i,y_i})}{\sum_{c=1}^{K} \exp(\eta_{i,c})}$
    \STATE \textbf{Update} network weights
\ENDFOR
\end{algorithmic}
\end{algorithm}

As depicted above, NoSAF can achieve high performance in deep model and heterophilic graph settings. However, continuous filtering may leads to a decrease in the total information contained within the nodes, resulting in almost complete disappearance of node representations at deeper layers \cite{van2020review}. To enhance network stability with increased depth, we propose a simple compensatory mechanism (cpm) $\mathcal{F}_{cpm}$, using the information in the codebank to compensate for the lost information:
\begin{equation}
    \mathcal{F}_{cpm}\left(\mathcal{C}^l, \gamma^l\right) = \mathcal{C}^{l} \circ (\mathbf{1} - {\mathcal B}( {{{\rm{\gamma }}^l}})   )
\end{equation}

Hence, the update formula for $H$ in NoSAF-D is expressed as following:
\begin{equation}
\begin{split}
    \quad H^{l+1} &= \mathcal{F}_f\left(Z^l, \gamma^l\right) + \mathcal{F}_{cpm}\left(C^l, \gamma^l\right) \\
                  &= \mathcal{Z}^{l} \circ {\mathcal B}( {{{\rm{\gamma }}^l}}) + \mathcal{C}^{l} \circ (\mathbf{1} - {\mathcal B}( {{{\rm{\gamma }}^l}}))
\end{split}
\end{equation}

\noindent NoSAF-D employs an information compensatory mechanism to further mitigate the issue of information loss during network deepening, ensuring the stability of training deep networks.

\vspace{0.6em}

\noindent \textbf{Theoretical Analysis.} In this section, we offer some analyses from Information Bottleneck (IB) perspective to demonstrate the effectiveness of NoSAF. Information Bottleneck (IB), which originates from the information theory, aims to find a compression code of the input signals while retaining as much valid information as possible from the original encoding \cite{tishby2000information, wu2020graph, yu2020graph, miao2022interpretable}.  Assuming the mapping function of NoSAF is denoted as ${\rm{\Psi }}$ and that of the traditional residual deepGNN as ${\rm{\Omega }}$, we can infer that ${\rm{I}}\left( {{\rm{\Psi }}\left( {\mathcal G} \right);{\rm{Y}}} \right) \ge {\rm{I}}\left( {{\rm{\Omega }}\left( {\mathcal G} \right);{\rm{Y}}} \right)$, indicating a higher information gain with NoSAF in mapping graph ${\mathcal G}$ to the output ${\rm{Y}}$. $I\left(  \cdot  \right)$ represents Shannon mutual information. 

As defined earlier, $Y = \left\{ {{{\rm{y}}_1},{{\rm{y}}_2}, \ldots, {{\rm{y}}_{\left| {\cal V} \right|}}} \right\}$ represents the labels of all nodes, and ${\cal G}$ is the input graph \cite{wang2022searching}. Methods based on GIB aim to identify an optimal subgraph ${\cal G}_s^*$ within the subgraph set $\mathbb{G}_{sub}\left( \cal{G} \right)$ by optimizing:
\begin{equation} \label{eq:GIB}
    \left. {max}_{\mathcal{G}_{sub} \in \mathbb{G}_{sub}{(\mathcal{G})}}I\left( {\mathcal{G}_{sub},Y} \right) - \beta I\left( {\mathcal{G}_{sub},\mathcal{G}} \right) \longrightarrow \mathcal{G}_{s}^{*} \right.
\end{equation}
\vspace{-1.0em}

\noindent In our framework, ${{\cal G}_{sub}}$ is identified as a specific subgraph within the larger graph ${\cal G}$. The notation $I\left( \cdot \right)$ is employed to represent Shannon mutual information, a crucial metric in our analysis. To effectively balance the dual aspects of our optimization equation, we introduce $\beta$ as a vital hyper-parameter. This parameter is strategically chosen to harmonize the intricate interplay between the essential elements of our graph-based model.

\textbf{Theoretical Analysis. Obs 1: There exist noise nodes at $l$-th layer (note nodes set as ${\dot {\cal G}_{sub}}^{l}$) which make no contributions to $I\left( {{\cal G},Y} \right)$ at all.} Considering that our target is to maximize $I\left( {{H^{L}},Y} \right)$. The mapping function ${\rm{\Psi }}$ is capable of filtering out the information from certain noisy nodes and their overly smoothed representations. This process effectively impedes the propagation of noise to the subsequent lower layers, enhancing the overall robustness and clarity of the network's output. However, the traditional residual deepGNN as ${\rm{\Omega }}$ can not achieve this target. With this in mind, we can obtain ${\rm{I}}\left( {{\rm{\Psi }}\left( {\mathcal G} \right);{\rm{Y}}} \right) > {\rm{I}}\left( {{\rm{\Omega }}\left( {\mathcal G} \right);{\rm{Y}}} \right)$.

\textbf{Theoretical Analysis. Obs 2: There does not exist noise nodes at $l$-th layer (note nodes set as ${\dot {\cal G}_{sub}}^{l}$) which make no contributions to $I\left( {{\cal G},Y} \right)$.} In such scenarios, NoSAF is capable of achieving effects comparable to traditional deep GNNs, without filtering out pertinent information.  This demonstrates NoSAF's effectiveness in maintaining the integrity of the data while simultaneously managing noise and over-smoothing issues, a crucial balance for deep learning models in graph-based applications. Then we can obtain ${\rm{I}}\left( {{\rm{\Psi }}\left( {\mathcal G} \right);{\rm{Y}}} \right) = {\rm{I}}\left( {{\rm{\Omega }}\left( {\mathcal G} \right);{\rm{Y}}} \right)$.

Based on \textbf{Obs 1} and \textbf{Obs 2}, we can summarize that ${\rm{I}}\left( {{\rm{\Psi }}\left( {\mathcal G} \right);{\rm{Y}}} \right) \ge {\rm{I}}\left( {{\rm{\Omega }}\left( {\mathcal G} \right);{\rm{Y}}} \right)$.

\section{Experiments}
Based on the model described, to validate our NoSAF, several critical research questions (\textbf{RQ}) need to be addressed. In this segment, our experiments are designed to answer the following research questions:

\begin{itemize}[leftmargin=*]
    \item \textbf{RQ1.} How do NoSAF and NoSAF-D compare in efficacy and performance ceilings to traditional shallow GNNs, deepGNNs and heterophilic GNNs?
    
    \item \textbf{RQ2.} Are the NoSAF and NoSAF-D models capable of scaling to large-scale graphs, which possess inherently complex internal structures and necessitate elevated resource commitments?

    \item \textbf{RQ3.} Is the NoSAF-D model adept at intensifying deep architectures to accommodate more challenging scenarios?
    \item \textbf{RQ4.} Does the adaptively learned node weights contribute meaningfully to the network's predictive capabilities?
    
\end{itemize}

To provide answers to these RQs, we orchestrate the following experiments:

\begin{itemize}[leftmargin=*]
    \item \textbf{Main experiment.} We conduct comparative analyses with mainstream models across a variety of graph datasets. The datasets utilized include nine homophilic graphs, six heterophilic graphs. The comparative study encompasses ten leading models.
    
    \item \textbf{Large-scale graph scalability experiment} In our large-scale graph scalability experiments, we compared our model with other advanced models on two large-scale graph datasets. This comparison was conducted to validate that our model not only scales effectively on large graphs but also maintains its leading-edge performance.
    
    \item \textbf{Depth scalability experiment.} We select two distinct datasets (a homophilic graph, a heterophilic graph) for our deep and scalability study. We compared the performance trends with other mainstream deep graph neural networks as the depth increased.
    
    \item \textbf{Ablation experiment.} In our ablation study, we select 6 graph benchmarks (three homophilic graphs, and three heterophilic graph). We explore the significance of \textbf{dynamically learned node weights} in the context of aggregating outputs across multiple layers. Our comparative analysis examines the performance impact of models employing this strategy versus those without it, while maintaining uniform parameter configurations.

\end{itemize}

\noindent Through these experiments, we anticipate drawing clear conclusions regarding the efficacy of NoSAF.

\subsection{Datasets \& Baselines}
\noindent \textbf{Baselines.} We evaluate the NoSAF model on a wide range of well-known benchmark graphs. Homophilic graphs including citation networks such as Cora, CiteSeer, and PubMed from \cite{yang2016revisiting}; the CoraFull dataset from \cite{bojchevski2017deep}; the coauthor networks CS and Physics from \cite{shchur2018pitfalls}; the Amazon networks Computers and Photo from \cite{shchur2018pitfalls}; the WikiCS dataset from \cite{mernyei2007wikipedia}. Heterophilic graphs including WebKB networks Cornell, Texas, and Wisconsin from \cite{pei2020geom}; Wikipedia-derived networks Chameleon and Squirrel from \cite{rozemberczki2021multi}; an actor co-occurrence network known as Actor from \cite{pei2020geom}. For large-scale graphs, we select ogbn-arxiv and ogbn-proteins from \cite{hu2020open}. Table \ref{tab:dataset_statistics} offers a comprehensive overview, showcasing the number of nodes, edges, features, classes, and the node homophily ratio, denoted as $\mathcal{H}$. These datasets encompass a broad spectrum, ranging from smaller graphs with approximately 200 nodes and 600 edges to larger ones exceeding 100,000 nodes and over 39 million edges. Among these datasets, there are instances of highly homophilic graphs, characterized by edge homophily values greater than 0.9, as well as heterophilic graphs, distinguished by homophily values less than 0.1.

\begin{table}[h]
\centering
\setlength{\tabcolsep}{4.5pt}
\caption{The statistics of the datasets.}
\label{tab:dataset_statistics}
\vspace{-0.3cm}
\begin{tabular}{lrrrrr}
\toprule
Dataset & \#Nodes & \#Edges & \#Features & \#Classes & $\mathcal{H}$ \\
\midrule
Cora & 2708 & 10556 & 1433 & 7 & 0.83 \\
CiteSeer & 3,327 & 9,104 & 3,703 & 6 & 0.71 \\
Pubmed & 19717 & 88648 & 500 & 3 & 0.79 \\
CoraFull & 19793 & 126842 & 8710 & 70 & 0.59 \\
Computers & 13752 & 491722 & 767 & 10 & 0.79 \\
Photo & 7650 & 238162 & 745 & 8 & 0.84 \\
CS & 18333 & 163788 & 6805 & 15 & 0.83 \\
Physics & 34493 & 495924 & 8414 & 5 & 0.92 \\
WikiCS & 11701 & 431726 & 300 & 10 & 0.66 \\
\hline
Cornell & 183 & 557 & 1,703 & 5 & 0.12 \\
Texas & 183 & 574 & 1,703 & 5 & 0.09 \\
Wisconsin & 251 & 916 & 1,703 & 5 & 0.17 \\
Chameleon & 2,277 & 62792 & 2,325 & 5 & 0.25 \\
Squirrel & 5,201 & 396846 & 2,089 & 5 & 0.22 \\
Actor & 7,600 & 53411 & 932 & 5 & 0.22 \\

\hline
ogbn-arxiv & 169,343 & 2,315,598 & 128 & 40 & 0.64\\
ogbn-proteins & 132,534 & 39,561,252 & 8 & - & - \\
\bottomrule
\end{tabular}
\end{table}

\noindent \textbf{Experiment Setup.} To ensure a fair comparison, for the OGBN datasets, we use the official splits provided by OGB\cite{hu2020open} and compare against independent models on the leaderboard that do not use additional data. For other datasets, we adapt a splitting strategy of 60\%, 20\%, and 20\% to divide the data into training, validation, and test sets, respectively. All models are run for 500 epochs, and the model with the highest accuracy on the validation set is selected as the best model. To mitigate randomness, all our experimental results are obtained by averaging \textbf{ten runs}. Our method is implemented using PyTorch Geometric, and all experiments are conducted on a single V100 32G GPU.

\noindent \textbf{Baselines.} In all the experiments we conducted, we selected ten models as baselines for comparison with our proposed model. These models include GCN \cite{kipf2016semi}, GAT \cite{velivckovic2017graph}, APPNP \cite{gasteiger2018predict}, MixHop \cite{abu2019mixhop}, H2GCN \cite{zhu2020beyond}, GPNN \cite{yang2022graph}, ResGCN \cite{li2019deepgcns}, JKNet \cite{xu2018representation}, GCNII \cite{chen2020simple}. These models encompass a wide range of GNN architectures, covering spectral-based GNNs, spatial-based GNNs, deep GNNs, as well as GNNs that limited the aggregation domain. This selection provides a comprehensive overview of the current state-of-the-art methodologies in the field, enabling a thorough evaluation of our model in comparison to established approaches.


\begin{table*}[t]  \small
\caption{Performance comparisons on different GNN architectures, in which we report best performance of these baselines. All experimental results are under \textbf{\underline{ten runs}}. * denotes tailed-made for heterophilic graphs. \textbf{\underline{The Geom-GCN results were taken from the original paper.}}}\label{tab:results1}
\vspace{-1.2em}
\setlength{\tabcolsep}{1.2pt}
\begin{center}
\def \arraystretch{0.94}
 \begin{tabular}{ccccccccccccc} 
 \toprule
    \multirow{2}{*}{\bf Backbone} & \multicolumn{6}{c}{\bf Shallow GNNs}  & \multicolumn{3}{c}{\bf Deep GNNs} & \multicolumn{2}{c}{\bf Ours}   \\

    \cmidrule(l){2-7} \cmidrule(l){8-10} \cmidrule(l){11-12}
    
     & \scriptsize \bf GCN  
     & \scriptsize \bf GAT  
     & \scriptsize \bf APPNP
     & \scriptsize \bf Mixhop*
     & \scriptsize \bf H2GCN*
     & \scriptsize \bf GPNN*
 
     & \scriptsize \bf JKNet 
     & \scriptsize \bf ResGCN 
     & \scriptsize \bf GCNII
     & \scriptsize \bf NoSAF 
     & \scriptsize \bf NoSAF-D  \\ 
     
     \midrule
     
       \multicolumn{11}{l}{\scriptsize{  \demph{ \it{homophilic graph benchmarks} }} }\\
       
        Cora  & $88.25_{\pm1.09}$  & $88.71_{\pm0.89}$  & 
        \cellcolor{gray!30}$89.30_{\pm1.45}$ & $88.08_{\pm1.12}$ & $87.14_{\pm1.35}$ & $86.20_{\pm1.55}$ & $87.84_{\pm1.44}$ & $87.10_{\pm1.44}$  & $88.52_{\pm1.40}$ & \underline{$88.87_{\pm1.14}$} & $88.08_{\pm1.52}$\\
        
        CiteSeer  & $77.14_{\pm1.17}$   & $76.39_{\pm1.27}$  & $76.80_{\pm1.10}$ & $76.96_{\pm1.13}$ & $76.42_{\pm1.25}$ & $76.20_{\pm1.82}$ & $75.68_{\pm1.88}$ & $76.06_{\pm2.12}$  & $77.05_{\pm0.78}$ & \cellcolor{gray!30}$78.05_{\pm1.70}$ & \underline{$77.31_{\pm1.38}$}\\
        
        PubMed & $89.01_{\pm0.41}$  & $88.00_{\pm0.60}$  & $89.72_{\pm0.56}$ & $90.16_{\pm0.53}$ & $88.54_{\pm0.47}$ &$88.72_{\pm0.56}$ & $88.66_{\pm0.23}$ & $89.53_{\pm0.34}$  & $90.14_{\pm0.46}$ & \underline{$90.50_{\pm0.47}$} & \cellcolor{gray!30}$90.62_{\pm0.47}$\\ 
        
        CoraFull & $71.85_{\pm0.74}$ & $70.60_{\pm0.89}$  & \underline{$71.88_{\pm0.76}$} &  $70.84_{\pm0.72}$ & $70.44_{\pm0.62}$ &$71.42_{\pm0.79}$ & $69.66_{\pm0.92}$ & $71.29_{\pm0.93}$ & $71.06_{\pm0.65}$ & \cellcolor{gray!30}$71.99_{\pm0.71}$ & $71.55_{\pm0.39}$\\ 

        Computers & $91.14_{\pm0.58}$ & $91.69_{\pm0.39}$ & $89.67_{\pm0.47}$ &  $92.03_{\pm0.47}$ & $91.53_{\pm0.64}$ &$90.28_{\pm0.52}$ & $91.51_{\pm0.38}$ & $91.96_{\pm0.53}$ & $89.77_{\pm0.49}$ & \underline{$92.09_{\pm0.68}$} & \cellcolor{gray!30}$92.45_{\pm0.49}$\\ 

        Photo & $93.93_{\pm0.62}$ & $94.55_{\pm0.53}$  & $95.06_{\pm0.30}$ &  $95.68_{\pm0.63}$ & $94.68_{\pm0.62}$ & $94.21_{\pm0.45}$ & $94.48_{\pm0.65}$ & $95.93_{\pm0.36}$ & $95.00_{\pm0.51}$ & \cellcolor{gray!30}$96.25_{\pm0.54}$ & \underline{$96.20_{\pm0.33}$}\\  

        CS & $93.77_{\pm0.37}$ & $93.27_{\pm0.30}$ & $95.91_{\pm0.23}$ &  $95.13_{\pm0.22}$ & $95.25_{\pm0.32}$  & $95.54_{\pm0.26}$& $94.08_{\pm0.37}$ & $95.62_{\pm0.42}$ & \underline{$96.03_{\pm0.27}$} & \cellcolor{gray!30}$96.09_{\pm0.27}$ & $95.94_{\pm0.33}$\\  

        Physics & $96.44_{\pm0.25}$ & $96.48_{\pm0.22}$ & $97.14_{\pm0.21}$ &  OOM  & $96.84_{\pm0.28}$ & $96.45_{\pm0.24}$ & $96.38_{\pm0.20}$ & $96.88_{\pm0.18}$ & $97.10_{\pm0.21}$ & \underline{$97.19_{\pm0.10}$} & \cellcolor{gray!30}$97.23_{\pm0.18}$\\ 

        WikiCS & $84.12_{\pm0.69}$ & $83.91_{\pm0.64}$ & $85.86_{\pm0.60}$ &  $85.82_{\pm0.84}$ & $85.22_{\pm0.76}$& $83.99_{\pm0.62}$& $84.17_{\pm0.76}$ & $84.73_{\pm0.74}$ & $85.10_{\pm0.65}$ & \underline{$86.06_{\pm0.61}$} & \cellcolor{gray!30}$86.21_{\pm0.47}$\\ 
    
    \midrule
           \multicolumn{11}{l}{\scriptsize{  \demph{ \it{heterophilic graph benchmarks} }} }\\
       
        Cornell & $51.62_{\pm7.59}$ & $50.00_{\pm7.57}$ & $77.30_{\pm6.64}$  &  $73.78_{\pm7.10}$ & $65.45_{\pm5.56}$ & $77.22_{\pm4.42}$ & $57.35_{\pm10.51}$ & $71.89_{\pm7.34}$ & $75.68_{\pm8.11}$ & \underline{$77.84_{\pm6.72}$} & \cellcolor{gray!30}$79.19_{\pm7.21}$\\ 

        Texas & $62.43_{\pm8.50}$ & $61.89_{\pm5.85}$ & $86.76_{\pm1.46}$ &  $81.35_{\pm9.31}$ & $87.14_{\pm6.50}$ & $84.22_{\pm6.44}$ & $63.19_{\pm3.70}$ & $82.70_{\pm6.77}$ & \cellcolor{gray!30}$87.84_{\pm4.40}$ & $85.41_{\pm6.14}$ & \underline{$87.57_{\pm7.77}$}\\ 
        
        Wisconsin & $56.20_{\pm5.76}$ & $56.40_{\pm5.99}$ & $86.00_{\pm3.10}$ &  $85.60_{\pm5.50}$ & $76.42_{\pm6.45}$ & $86.42_{\pm3.28}$ & $53.60_{\pm10.28}$ & $83.40_{\pm6.47}$ & $88.00_{\pm4.29}$ & \underline{$88.20_{\pm6.63}$} & \cellcolor{gray!30}$89.40_{\pm2.50}$\\ 

        Chameleon & $68.75_{\pm2.27}$ & $66.07_{\pm1.85}$  & $68.07_{\pm1.70}$ &  $67.27_{\pm3.25}$ & $65.92_{\pm1.88}$ & \underline{$ 69.14_{\pm1.69}$} & $65.14_{\pm1.69}$ & $67.67_{\pm1.89}$ & $67.58_{\pm1.35}$ & \cellcolor{gray!30}$70.31_{\pm1.74}$ & \underline{$69.14_{\pm3.14}$}\\ 

        Squirrel & $55.66_{\pm1.68}$ & $56.00_{\pm2.59}$ & $54.17_{\pm2.22}$ &  $51.18_{\pm1.39}$ & $53.18_{\pm1.96}$ &$59.04_{\pm1.26}$ & $52.26_{\pm1.77}$ & $59.58_{\pm2.23}$ & $55.13_{\pm2.70}$ & \underline{$60.25_{\pm0.99}$} & \cellcolor{gray!30}$61.23_{\pm1.26}$\\  

        Actor & $30.43_{\pm1.59}$ & $29.26_{\pm0.91}$ & $37.05_{\pm1.17}$ &  $36.09_{\pm1.58}$  & $36.58_{\pm1.25}$ & $37.11_{\pm1.24}$& $27.99_{\pm1.29}$ & $35.82_{\pm1.25}$ & $37.66_{\pm1.10}$ & \underline{$38.56_{\pm1.44}$} & \cellcolor{gray!30}$38.76_{\pm1.14}$\\

    \midrule
\end{tabular} \label{main1}
\end{center}
\end{table*}

\subsection{Main Results (RQ1)} \label{RQ1}

In this section, we compare the efficiency of NoSAF(-D) with previous shallow and deep GNN methods from two perspectives: \textbf{(I)} optimal performance across different benchmarks; \textbf{(II)} capacity of handling complex heterophilic datasets. To better compare model performances, we replicate all baselines using the optimal parameters provided in the literature. All outcomes are presented as the mean and variance of ten iterations. As shown in Table \ref{main1}, we highlight the top-performing model by shading its background in gray and underscore the second-best with an underline. We list the observations as follow:


\textbf{Obs.1.} \textbf{Across 15 graph benchmark tasks, NoSAF (and -D version) achieve optimal performance in 13 set of them.} \underline{Specifically}, they secure the first-place performance in 8 out of 9 homophilic graphs and in 5 out of 6 heterophilic graphs. Compared to deep models based on skip connections, such as JKNet, ResGCN, which typically excel at handling homophilic graph settings, our models even demonstrate a clear superiority in performance. \underline{Further,} across all homophilic graphs, the best performances are consistently achieved by our models. Although some shallow architectures exhibit exceptional performance on small datasets, our model continues to outperform them in the vast majority of homophilic settings. Specially, in the WikiCS dataset ($\mathcal{H}=0.66$), NoSAF shows significant improvement over these models. Compared to the best performing model among them, GCNII (85.10\%), our NoSAF model improves by $0.96\% \uparrow$ and the NoSAF-D model improves by $1.11\% \uparrow$.

\textbf{Obs.2.} \textbf{On complex heterophilic datasets ($\mathcal{H}<0.50$), NoSAF (and -D version) still exhibit exceptional performance, demonstrating no significant sensitivity to data variations.} On the Cornell, Wisconsin, Chameleon, Squirrel, and Actor datasets, our proposed two models clinch both first and second place. Our models demonstrate improvements of over 1.0\% relative to the best results of other baselines: 1.89\% in Cornell, 1.40\% in Wisconsin, 1.17\% in Chameleon, 1.65\% in Squirrel, and 1.10\% in Actor. 

\textbf{Summary.} Based on \textbf{Obs.1.} and \textbf{Obs.2.}, we systematically validate the upper bound of the model's performance, providing evidence for the potential of NoSAF as a universal operator.

\subsection{Large-scale benchmarks scalability (RQ2)} \label{RQ2}

Due to the high structural diversity of large graphs, their characteristics may more closely align with real-world data. This complexity and variety in structure often mean that many GNNs. In this section, we apply NoSAF and NoSAF-D to large graphs with numerous nodes and dense connections. We select two datasets, Ogbn-Arxiv ($\#nodes=169,343; \#edges=1,166,243$) and Ogbn-proteins ($\#nodes=132,534; \#edges=39,561,252$), to assess the model's scalability on large-scale graph benchmarks. Concretely, for Ogbn-Arxiv, we select GNN frameworks GCN, DeeperGCN \cite{li2020deepergcn}, DAGNN \cite{liu2020towards}, JKNet, GCNII, GTAN \cite{wu2022gtnet}, UniMP \cite{shi2020masked}, LEGNN \cite{yu2022labelenhanced} and AGDN \cite{sun2022adaptive} as the comparative models. For Ogbn-Protein, we choose GCN, GraphSAGE, DeepGCN, DeeperGCN, UniMP, RevGNN-Wide (Deep) \cite{li2020deepergcn}, AGDN and GIPA \cite{li2023gipa} as base models \footnote{These models are sourced from the OGB Leaderboard \url{https://ogb.stanford.edu/docs/leader_nodeprop/}. Worth noting that instead of using hybrid models, we have chosen the best-performing single models for comparison to ensure fairness.}. 

\begin{figure}[h]
\centering
\includegraphics[width=1.0\columnwidth]{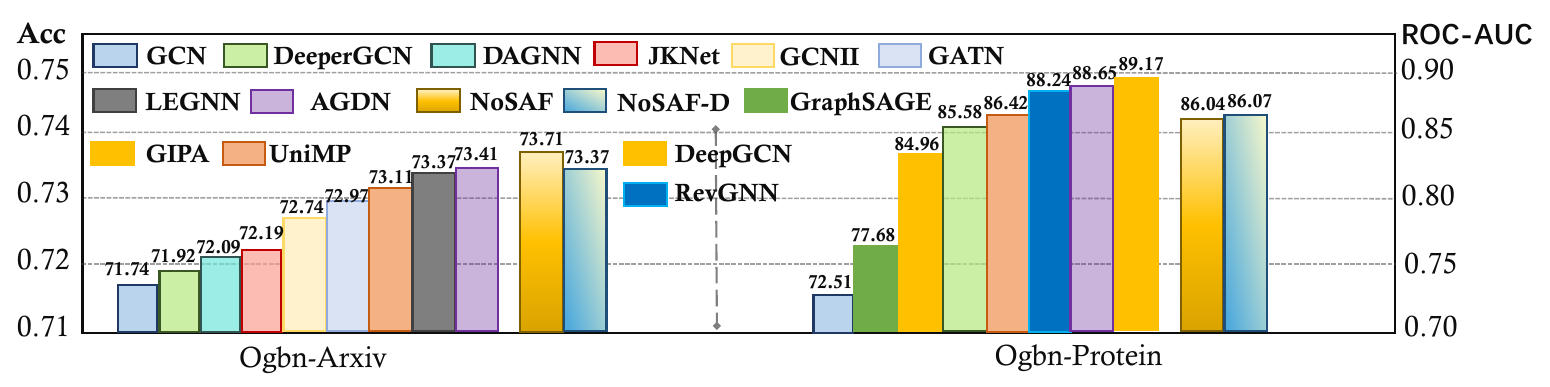}
\caption{Model performance across large-scale graphs.}
 \vspace{-0.4cm}
\label{fig:rq2fig}
\end{figure}

\noindent We list the observations \textbf{(Obs 3.)} that on the Arxiv benchmark task, the NoSAF model achieved state-of-the-art (SOTA) performance across all backbones, surpassing the second-best model by 0.3\% in accuracy. Simultaneously, the NoSAF-D model also secured a top-3 performance position. On the Proteins task, we employed GEN \cite{li2020deepergcn} as our GNN block, identical to that used in DeeperGCN. It is observed that, compared to DeeperGCN, our two models achieved performance improvements of 0.24\% and 0.27\%, respectively. However, due to limitations inherent in GEN, they did not reach very high performance levels.

\begin{table*}[h] 
\footnotesize
\setlength{\tabcolsep}{10pt}
  \caption{Comparison results among different benchmarks across differnet layer settings.} 
  \vspace{-0.2cm}
  \centering
  \scalebox{0.92}{
  \begin{tabular}{cccccccc}
    \toprule
     \multirow{2}{*}{\makecell{\textbf{Benchmark}\\PubMed $\mathcal{H}=0.79$}} &  \multicolumn{3}{c}{\textbf{Shallow-layers}}  &  \multicolumn{3}{c}{\textbf{Deep-layers}}  \\
     
     \cmidrule(lr){2-4} \cmidrule{5-7} 
     
    & \textbf{2-L} &  \textbf{4-L}  &  \textbf{8-L} & \textbf{16-L} &  \textbf{32-L}  &  \textbf{64-L}  \\
    \midrule
    JKNet   & ${88.36_{\pm0.50}}$ & ${88.61_{\pm  0.38}}$ & ${88.55_{\pm 0.43}}$  & ${88.47_{\pm0.59}}$ &\cellcolor{gray!30}${88.62_{\pm 0.52}}$ & ${88.54_{\pm 0.41}}$ \\
    
    ResGCN  & ${89.02_{\pm0.34}}$  & $\cellcolor{gray!30}{89.24_{\pm0.47}}$ & ${89.06_{\pm0.31}}$ & ${88.84_{\pm0.55}}$ & ${88.55_{\pm0.40}}$ & ${88.58_{\pm0.31}}$ \\
      
    GCNII   & $\cellcolor{gray!30}{90.14_{\pm0.46}}$  & ${90.08_{\pm0.38}}$ & ${90.02_{\pm0.38}}$ & ${89.76_{\pm0.39}}$ & ${89.97_{\pm0.20}}$ & ${89.98_{\pm0.43}}$   \\

    NoSAF   & ${90.23_{\pm0.65}}$  & ${90.32_{\pm0.39}}$ & \cellcolor{gray!30}${90.35_{\pm0.26}}$ & ${90.32_{\pm0.40}}$ & ${90.32_{\pm0.37}}$ & ${88.32_{\pm  1.50}}$  \\

    NoSAF-D & ${90.23_{\pm0.71}}$  & ${90.25_{\pm0.52}}$ & ${90.41_{\pm0.27}}$ & ${90.29_{\pm0.46}}$ & $\cellcolor{gray!30}{90.45_{\pm0.43}}$ & ${90.37_{\pm  0.52}}$    \\

    \midrule
    \multirow{2}{*}{\makecell{\textbf{Benchmark}\\Actor $\mathcal{H}=0.22$}}  &  \multicolumn{3}{c}{\textbf{Shallow-layers}}  &  \multicolumn{3}{c}{\textbf{Deep-layers}}  \\
     
     \cmidrule(lr){2-4} \cmidrule{5-7} 
     
    & \textbf{2-L} &  \textbf{4-L}  &  \textbf{8-L} & \textbf{16-L} &  \textbf{32-L}  &  \textbf{64-L}  \\
    \midrule
    JKNet   & ${28.04_{\pm 2.07}}$ & ${28.67_{\pm1.32}}$ & ${28.91_{\pm 1.06}}$  & ${28.38_{\pm1.38}}$ &$\cellcolor{gray!30}{29.86_{\pm1.53}}$ & ${29.48_{\pm 1.30}}$ \\
    
    ResGCN  & ${34.66_{\pm1.84}}$  & $\cellcolor{gray!30}{35.82_{\pm1.25}}$ & ${34.55_{\pm0.88}}$ & ${35.00_{\pm1.69}}$ & ${34.61_{\pm0.81}}$ & ${33.98_{\pm1.10}}$ \\
      
    GCNII   & ${37.49_{\pm1.04}}$  & ${37.52_{\pm0.90}}$ & ${37.56_{\pm1.48}}$ & ${37.53_{\pm1.15}}$ & ${37.30_{\pm0.88}}$ & $\cellcolor{gray!30}{37.66_{\pm0.64}}$ \\

    NoSAF   & $\cellcolor{gray!30}{38.64_{\pm1.21}}$  & ${38.07_{\pm1.53}}$ & ${38.43_{\pm1.24}}$ & ${38.50_{\pm1.64}}$ & ${38.24_{\pm0.92}}$ & ${38.25_{\pm1.52}}$  \\

    NoSAF-D & ${38.20_{\pm1.40}}$  & ${37.84_{\pm1.27}}$ & ${38.18_{\pm0.75}}$ & ${37.91_{\pm0.73}}$ & ${38.22_{\pm1.19}}$ & $\cellcolor{gray!30}{38.56_{\pm1.28}}$  \\
    \bottomrule
  \end{tabular}\label{tab:rq2}
  }
  \vspace{-1em}
\end{table*}

\vspace{-0.3cm}
\subsection{Depth scalability experiments (RQ3)}

In this section, we explore the capabilities of NoSAF and NoSAF-D in deep network architectures. For our benchmarks, we select a homophilic graph (PubMed) and a heterophilic graph (Actor). We can list the observations as follow:
\begin{figure}[h]
\centering
\includegraphics[width=1.0\columnwidth]{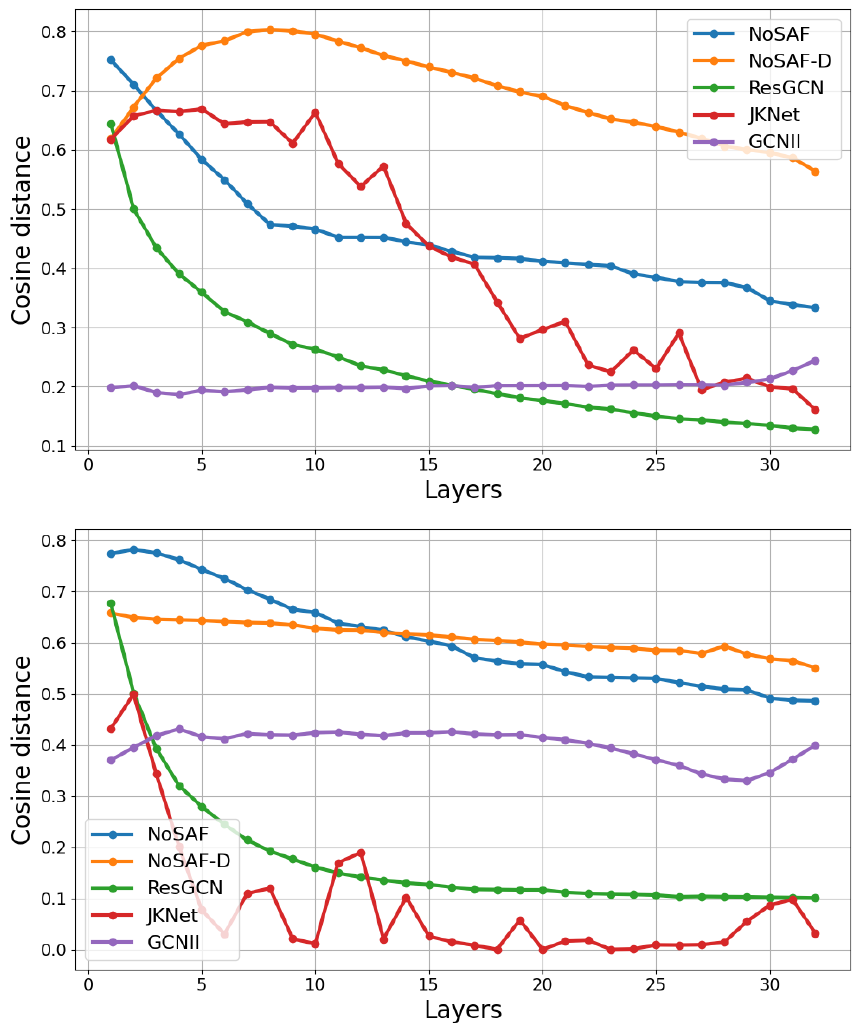}
\vspace{-0.5cm}
\caption{The comparison of the smoothness of the output at each layer for all deep models. The line graph on the upper side displays the results on PubMed (\(\mathcal{H}=0.79\)), while the line graph on the lower side shows the results on Actor (\(\mathcal{H}=0.22\)).}
 \vspace{-0.2cm}
\label{fig:smoothness}
\end{figure}

\noindent \textbf{Obs.4.} \textbf{NoSAF can achieve higher performance across homophilic or heterophilic graphs.}  As shown in Tab \ref{tab:rq2}, it is observable that NoSAF and NoSAF-D consistently outperforms other deep GNN methods across all different layers benchmarks, attaining improvements in homophilic graph performance gains from 1.64\% to 1.83\% and in heterophilic graph performance gains about $\sim1.00\%-10.00\%$. Specifically, on NoSAF-D + 64-layer setting, NoSAF achieves a performance close to 38.56\%, surpassing the previously strong contenders, JKNet at 29.48\% and GCNII at 37.66\%, by a large margin.

To more thoroughly investigate the model's effect on mitigating oversmoothing, we compare the smoothness of the output from each layer of the NoSAF and NoSAF-D models with that of other deepGNNs, as shown in Figure \ref{fig:smoothness}. We measure the smoothness using the average cosine distance \(D_{avg}\) among all nodes, where a smaller \(D_{avg}\) indicates greater similarity between all nodes. For the node features \(H^{l} = \{H_{1}^l, H_{2}^l, \ldots, H_{N}^l\}\) at the \(l\)-th layer, where \(N\) denotes the node features, the calculation for the average cosine distance \(D^{l}_{avg}\) at layer \(l\) is as follows:

\begin{equation}
    S^{l}_{ij} = \frac{H^{l}_{i} \cdot H^{l}_j}{\|H^{l}_{i}\| \cdot \|H^{l}_j\|}
\end{equation}
    
\begin{equation}
    D^{l}_{ij} = 1 - S^{l}_{ij}
\end{equation}

\begin{equation}
    D^{l}_{avg} = \frac{2}{N(N-1)} \sum_{i=1}^N \sum_{j=i+1}^N Dij
\end{equation}

Where \( S^{l}_{ij} \) represents the cosine similarity between the features of nodes \( i \) and \( j \) at the \( l \)-th layer.From Figure \ref{fig:smoothness}, it is evident that our method, especially NoSAF-D, is more effective in mitigating oversmoothing compared to ResGCN and JKNet, achieving significant suppression of oversmoothing in both homophilic and heterophilic graphs. Interestingly, GCNII did not even exhibit signs of smoothing, which we believe might be due to the continuous input of information from the first layer to subsequent layers. However, appropriate intra-class smoothing is crucial for handling node classification tasks, which is also why our model outperforms GCNII.

\begin{figure}[h]
\centering
\includegraphics[width=1.0\columnwidth]{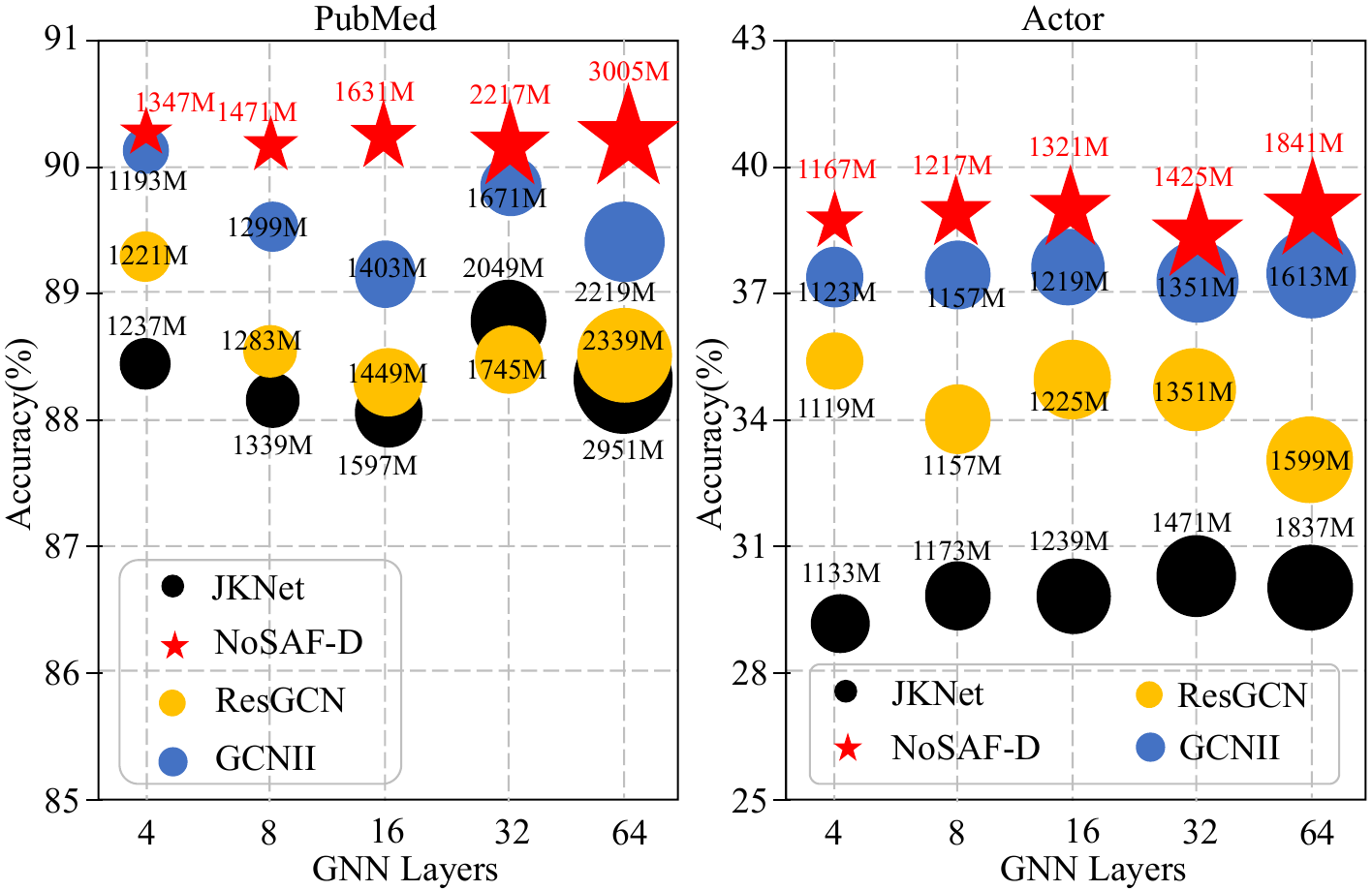}
\vspace{-0.5cm}
\caption{Summary of our achieved performance (y-axis) at different graph and GNN layers (x-axis) on PubMed and Actor. The size of markers represent the memory usage. Red stars (\(\textcolor{red}{\star}\)) are established by our method.}
 \vspace{-0.2cm}
\label{fig:rq2}
\end{figure}

\noindent \textbf{Obs.5. NoSAF and its -D version do not incur a significant storage overhead.} Unlike complex deep GNN methods, which typically experience a significant decline in performance with increased memory usage, NoSAF improves model performance without imposing a serious storage burden. As shown in Figure \ref{fig:rq2}, NoSAF's node filtering approach and codebank strategy introduce only about 1/10 to 1/3 of the storage pressure. 


\subsection{Ablation experiments (RQ4)}
In this section, we explore the role of each component in the NoSAF model. We design two experiments to support our ablation study: The first experiment involves progressively removing components from the NoSAF-D model and evaluating the performance of the resulting models. For this experiment, we fix the model depth at 16 layers and the number of neurons in the hidden layer at 64. Except for the components that are removed, the rest of the model's structure remains unchanged. The second experiment assesses the performance of all models from the first experiment at various layer depths, to verify our proposed structure's ability to efficiently scale the model to greater depths. We select PubMed ($\mathcal{H}=0.79$), WikiCS ($\mathcal{H}=0.66$), CoraFull ($\mathcal{H}=0.59$), Squirrel ($\mathcal{H}=0.22$), Actor ($\mathcal{H}=0.22$), and Wisconsin ($\mathcal{H}=0.17$) for evaluating the models. These datasets exhibit a gradual decline in homophily, from 0.79 to 0.17. This allows us to demonstrate the effectiveness of the models in our experiment across datasets with varying characteristics.

\begin{table}[h]
    \centering
    \caption{Ablation results across different settings. ``cpm'',  ``nw'' and ``cb'' denote compensatory mechanism, node weight and codebank.} 
    \vspace{-1.0em}
     \label{tab:human}
     \resizebox{0.48\textwidth}{!}{ 
    \begin{tabular}{ccccc}
    \toprule 
    \multirow{2}{*}{ Model } & \multicolumn{2}{c}{Original} & \multicolumn{2}{c}{Designs} \\ 
    \cmidrule(l){2-5}
    & NoSAF-D & w/o cmp  & w/o cmp+nw & w/o cmp+nw+cb \\ 
    \midrule
PubMed        & ${90.33_{\pm0.30}}$ & ${90.31_{\pm0.27}}(\downarrow0.02)$ & ${90.22_{\pm0.32}}(\downarrow0.09)$ & ${86.29_{\pm0.61}}(\downarrow3.93)$ \\
WikiCS        & ${85.53_{\pm0.59}}$ & ${85.42_{\pm0.47}}(\downarrow0.11)$ & ${85.30_{\pm0.58}}(\downarrow0.12)$ & ${81.32_{\pm0.74}}(\downarrow3.98)$ \\
CoraFull      & ${71.20_{\pm0.57}}$ & ${70.67_{\pm0.57}}(\downarrow0.53)$ & ${70.43_{\pm0.67}}(\downarrow0.24)$ & ${65.65_{\pm0.47}}(\downarrow4.78)$ \\
Squirrel      & ${61.23_{\pm1.22}}$ & ${60.35_{\pm1.28}}(\downarrow0.88)$ & ${58.31_{\pm1.54}}(\downarrow2.04)$ & ${59.30_{\pm1.75}}(\uparrow0.99)$ \\
Actor         & ${37.91_{\pm0.73}}$ & ${38.50_{\pm1.64}}(\uparrow0.59)$ & ${37.81_{\pm1.32}}(\downarrow0.69)$ & ${25.84_{\pm1.20}}(\downarrow11.97)$ \\
Wisconsin     & ${87.20_{\pm3.43}}$ & ${86.00_{\pm4.81}}(\downarrow1.20)$ & ${83.20_{\pm4.13}}(\downarrow2.80)$ & ${56.40_{\pm4.20}}(\downarrow26.8)$ \\
    \bottomrule
    \end{tabular}}
\end{table}

\noindent \textbf{Obs 6.} From Table 3, we observe that the model's performance drops most significantly when the codebank is removed. The presence of the codebank facilitates the easy back-propagation of gradients to every layer of the model, which is key to deepening our model. At a depth of 16 layers, the stacking of GCN blocks without a codebank makes it challenging to train the model effectively. In models with a codebank, our node weight structure proves to be extremely important in heterophilic scenarios. We can observe that removing the node weight structure leads to a performance decrease of 2.04 on the Squirrel dataset and 2.80 on the Actor dataset. In homophilic scenarios, although the node weight structure plays a role, its impact is not as pronounced as in heterophilic scenarios.

\begin{figure}[t]
  \centering
  \includegraphics[width=1\linewidth]{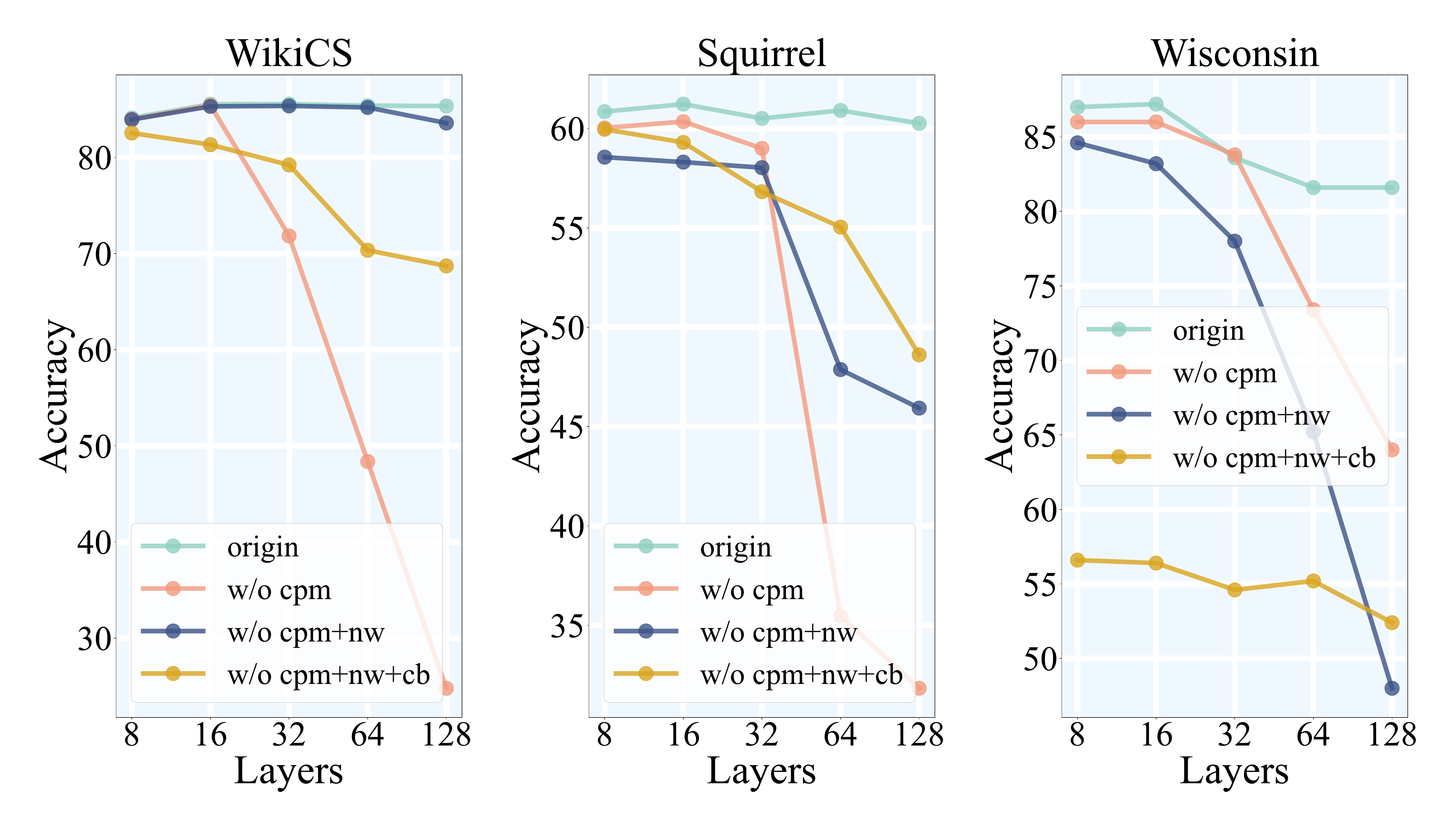}
  \vspace{-0.6cm}
  \caption{Performance (y-axis) of the NoSAF-D model and its ablation variants without cpm, without cpm+nw, and without cpm+nw+cb across different GNN layers (x-axis) on the WikiCS, Squirrel, and Wisconsin datasets.}
  \label{fig:RQ4}
\end{figure}

\vspace{0.3em}
\noindent \textbf{Obs 7.} As shown in Fig \ref{fig:RQ4}, with the number of layers increases, the effectiveness of our compensatory mechanism becomes increasingly prominent. NoSAF-D is capable of learning robust node representations even at ultra-depths. \textbf{On WikiCS and Squirrel, the model shows no performance decline even at 128 layers.} In contrast, the performance of NoSAF, which lacks this compensatory mechanism, often starts to decline significantly after 32 layers. An interesting observation from the figure is that, without the filtering mechanism, the model can still reach great depths in homophilic scenarios but not in heterophilic ones, which suggests that fully trusting information aggregated by neighborhood aggregation strategies in heterophilic graphs is not a wise choice. Especially when the model is very deep, it may accumulate excessive noise. Our filtering strategy effectively alleviates this issue.

\vspace{-0.2cm}

\section{Conclusion}
Recognizing that current Deep GNNs still rely on GNNs designed based on the homophily assumption as their core blocks, in this paper, we introduce NoSAF and its deep version, NoSAF-D. NoSAF sequentially filters the aggregated information of each node at every layer and integrates the filtered information into the codebank, which ensures that the model can learn effective node representations even in heterophilic graph context. To address the substantial information loss caused by the filtering strategy as model depth increases, NoSAF-D utilizes the information in the codebank to supplement the filtered data, thereby preventing performance degradation with added depth. Our experiments demonstrate that our designs achieves SOTAs in many graph benchmark tasks and maintains its performance as the number of layers increases. 

\section{Broader Impact \& Further Work}

This work is part of a broader line of research aimed at deepening the network depth of state-of-the-art models in Graph Neural Networks (GNNs). This line of research holds the potential for positive societal impacts, stemming from the scaling of GNN models and their practical implementations. In the future, we will pay closer attention to the design for heterogeneous and homophilic graphs, particularly in conjunction with mainstream heterogeneous GNNs, and enhance the generalization capabilities of our framework design.

\section{Acknowledgement}
This paper is partially supported by the National Natural Science Foundation of China (No.62072427, No.12227901), the Project of Stable Support for Youth Team in Basic Research Field, CAS (No.YSBR-005), Academic Leaders Cultivation Program, USTC.

\bibliographystyle{IEEEtran}
{
\bibliography{sample-base}
}

\begin{IEEEbiography}
[{\includegraphics[width=1in,height=1.25in,clip,keepaspectratio]{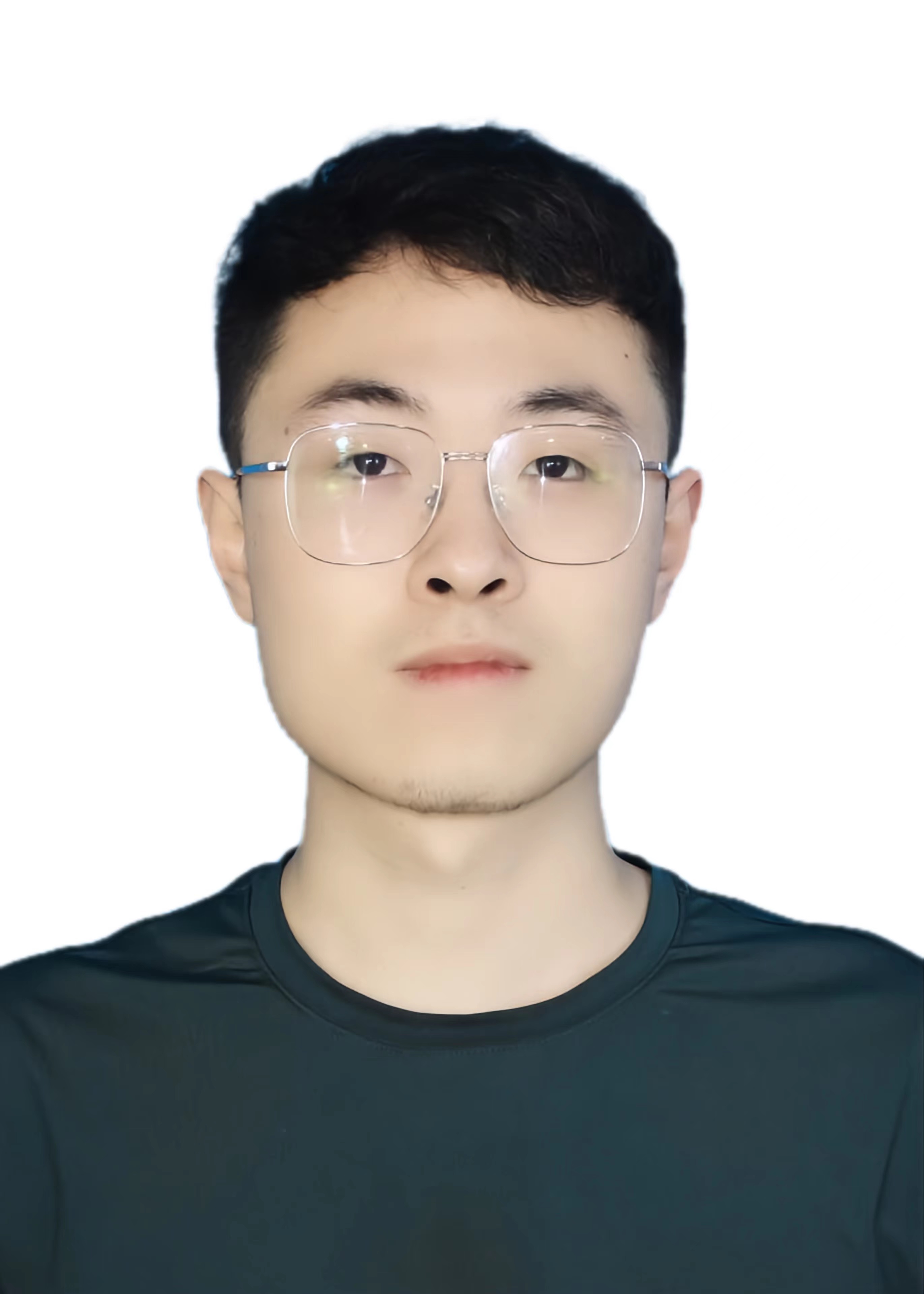}}]{Shilong Wang} is currently a master student in the Department of Computer Science and Technology, University of Science and Technology of China (USTC). His research interests are Graph Neural Networks and data centric AI.
\end{IEEEbiography}
\vspace{-1.2cm}

\begin{IEEEbiography}
[{\includegraphics[width=1in,height=1.25in,clip,keepaspectratio]{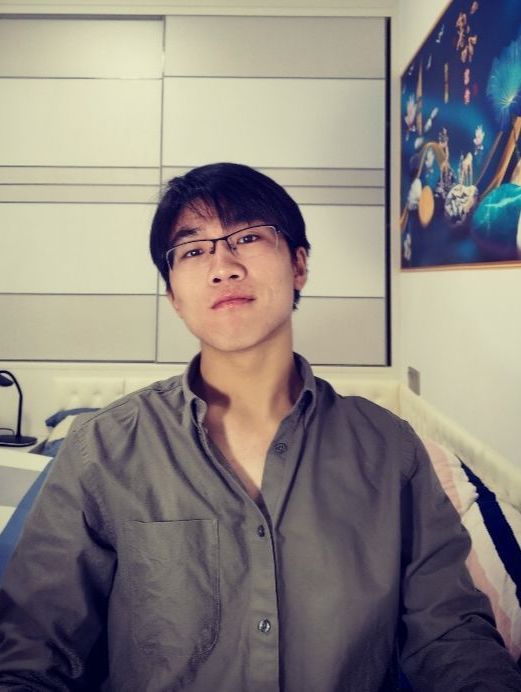}}]{Hao Wu} is a master's student who is currently undergoing joint training by the Department of Computer Science and Technology at the University of Science and Technology of China (USTC) and the Machine Learning Platform Department at Tencent TEG. His research interests encompass various areas, including spatio-temporal data mining, modeling of physical dynamical systems, and meta-learning, among others. He has published papers on top-tier conferences and journal such as ICML, NeurIPS, ICLR, AAAI and TKDE.
\end{IEEEbiography}
\vspace{-1.2cm}

\begin{IEEEbiography}
[{\includegraphics[width=1in,height=1.25in,clip,keepaspectratio]{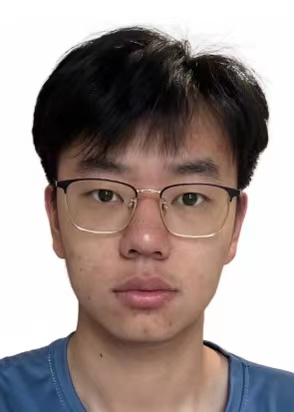}}]{Yifan Duan} is currently a master student in the Department of Software Engineering, University of Science andTechnology of China (USTC). His research interests are data mining and Graph Neural Networks.
\end{IEEEbiography}
\vspace{-1.2cm}

\begin{IEEEbiography}
[{\includegraphics[width=1in,height=1.25in,clip,keepaspectratio]{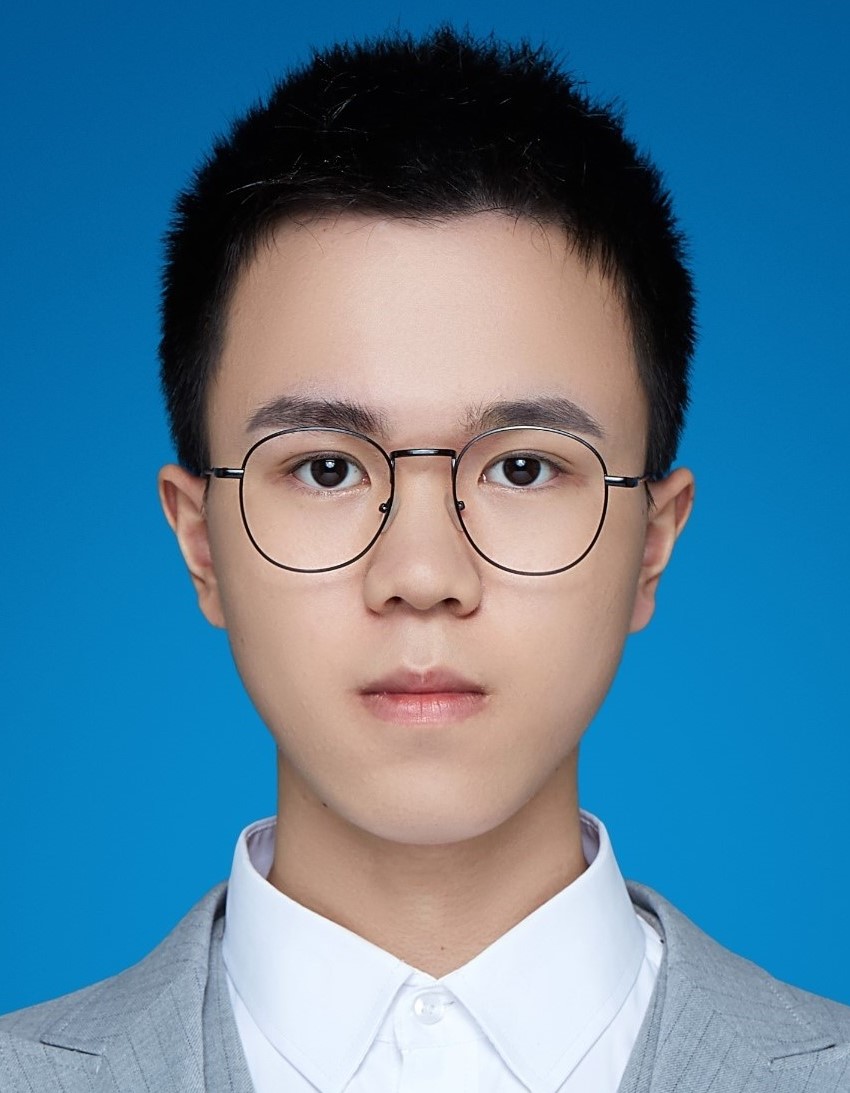}}]{Guibin Zhang} is currently an undergraduate in the Department of Computer Science and Technology, College of Electronic and Information Engineering, major in Data Science, Tongji University, Shanghai, China. His research interest include data mining, graph representation learning and knowledge modeling.
\end{IEEEbiography}
\vspace{-1.2cm}

\begin{IEEEbiography}[{\includegraphics[width=1in,height=1.25in,clip,keepaspectratio]{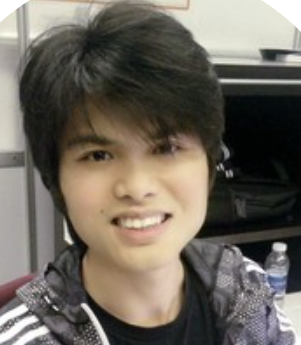}}]{Guohao Li} obtained his BEng degree in Communication Engineering from Harbin Institute of Technology in 2015. In 2018, he received his Master Degree in Communication and Information Systems from Chinese Academy of Science. He was a research intern at SenseTime and Intel ISL. He received his PhD degree from King Abdullah University of Science and Technology. His primary research interests lie in the fields of Computer Vision, Robotics, and Deep Learning.
\end{IEEEbiography}
\vspace{-1.2cm}

\begin{IEEEbiography}[{\includegraphics[width=1in,height=1.25in,clip,keepaspectratio]{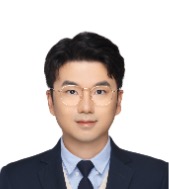}}]{Yuxuan Liang} (Member, IEEE) is currently an Assistant Professor at Intelligent Transportation Thrust, also affiliated with Data Science and Analytics Thrust, Hong Kong University of Science and Technology (Guangzhou). He is working on the research, development, and innovation of spatio-temporal data mining and AI, with a broad range of applications in smart cities. Prior to that, he completed his PhD study at NUS. He published over 50 peer-reviewed papers in refereed journals and conferences, such as TKDE, AI Journal, TMC, KDD, WWW, NeurIPS, and ICLR. He was recognized as 1 out of 10 most innovative and impactful PhD students focusing on data science in Singapore by Singapore Data Science Consortium.
\end{IEEEbiography}
\vspace{-1.2cm}

\begin{IEEEbiography}[{\includegraphics[width=1in,height=1.25in,clip,keepaspectratio]{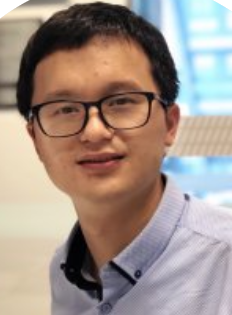}}]{Shirui Pan} (Senior Member, IEEE) received the Ph.D. degree in computer science from the University of Technology Sydney (UTS), Ultimo, NSW, Australia, in 2015. 

He is currently a Professor with the School of Information and Communication Technology Griffith University, Brisbane, QLD, Australia. Prior to this, he was a Senior Lecturer with the Faculty of IT at Monash University, Melbourne, VIC, Australia. His research interests include data mining and machine learning. To date, he has published over 100 research papers in top-tier journals and conferences, including IEEE TRANSACTIONS ON PATTERN ANALYSIS AND MACHINE INTELLIGENCE, IEEE TRANSACTIONS ON KNOWLEDGE AND DATA ENGINEERING. IEEE TRANSACTIONS ON NEURAL NETWORKS AND LEARNING SYSTEMS, ICML, NeurlPS, and KDD. 

Dr. Pan is an ARC Future Fellow and a fellow of the Queensland Academy of Arts and Sciences (FQA). His research received the 2024 CIS IEEE TRANSACTIONS ON NEURAL NETWORKSAND LEARNING SYSTEMs Outstanding Paper Award and the 2020 lEEE ICDM Best Student Paper Award. He is recognized as one of the AI 2000 AAAl/ICAl Most Infuential Scholars in Australia.
\end{IEEEbiography}
\vspace{-1.2cm}

\begin{IEEEbiography}[{\includegraphics[width=1in,height=1.25in,clip,keepaspectratio]{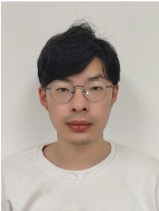}}]{Kun Wang} is currently a Ph.D. candidate of University of Science and Technology of China (USTC). He has published papers on top conferences and journals such as ICLR and TKDE. His primary research interests are Sparse Neural Networks, ML4Science and Graph Neural Networks.
\end{IEEEbiography}
\vspace{-1.2cm}

\begin{IEEEbiography}[{\includegraphics[width=1in,height=1.25in,clip,keepaspectratio]{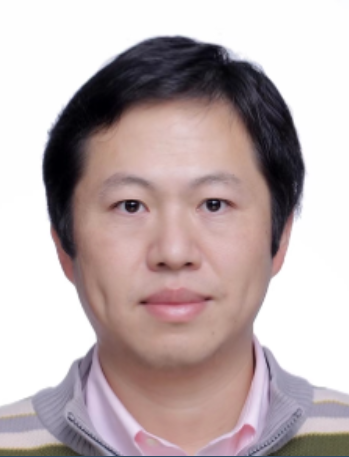}}]{Yang Wang} is now an associate professor at School of Computer Science and Technology, School of Software Engineering, and School of Data Science in USTC. He got his Ph.D. degree at University of Science and Technology of China in 2007. Since then, he keeps working at USTC till now as a postdoc and an associate professor successively. Meanwhile, he also serves as the vice dean of school of software engineering of USTC. His research interest mainly includes wireless (sensor) networks, distribute systems, data mining, and machine learning, and he is also interested in all kinds of applications of AI and data mining technologies especially in urban computing and AI4Science.
\end{IEEEbiography}
\vspace{-1.2cm}





\ifCLASSOPTIONcaptionsoff
  \newpage
\fi

\end{document}